%% file: main.tex
\documentclass[11pt]{article}

\newcommand{\paperversion}{preprint}
\expandafter\usepackage\expandafter[\paperversion]{acl}
\makeatletter
\let\ifreviewversion\ifacl@anonymize
\makeatother
\ifreviewversion
  \newcommand{\repourl}{https://github.com/metacognitive-behavioral-tuning/metacognitive-behavioral-tuning}
\else
  \newcommand{\repourl}{https://github.com/12kimih/metacognitive-behavioral-tuning}
\fi
\ifreviewversion
\ifdefined\pdfinfoomitdate \pdfinfoomitdate=1 \pdftrailerid{} \fi
\fi

\usepackage{times}
\usepackage{latexsym}
\usepackage[T1]{fontenc}
\usepackage[utf8]{inputenc}
\usepackage{microtype}
\usepackage{inconsolata}
\usepackage{graphicx}
\usepackage{xurl}

\usepackage{amsmath}
\usepackage{amssymb}
\usepackage{booktabs}
\usepackage{tabularx}
\usepackage{array}
\usepackage{makecell}
\usepackage{enumitem}
\usepackage{xcolor}
\usepackage{colortbl}

\input{macros}
\input{tables/numbers}

\title{Distilling Metacognitive Behaviors through Agentic Editing}

\author{%
  Ik-hwan Kim\textsuperscript{*} \\
  Dept.\ of ECE\\
  Seoul Nat'l Univ.\\
  \And
  Hyeongrok Han\textsuperscript{*} \\
  Dept.\ of ECE\\
  Seoul Nat'l Univ.\\
  \And
  Mingi Jung \\
  Dept.\ of ECE\\
  Seoul Nat'l Univ.\\
  \And
  Sangwon Yu\textsuperscript{\ddag} \\
  Dept.\ of ECE\\
  Seoul Nat'l Univ.\\
  \AND
  Jinseok Hong \\
  AX/PI Center\\
  Samsung Elec.\\
  \And
  Sang Hun Kim \\
  AX/PI Center\\
  Samsung Elec.\\
  \And
  Yoonyoung Choi \\
  AX/PI Center\\
  Samsung Elec.\\
  \And
  Sungroh Yoon\textsuperscript{\dag} \\
  ECE and AI\\
  Seoul Nat'l Univ.\\
}

\begin{document}
\maketitle
\ifreviewversion\else
\renewcommand{\thefootnote}{}
\footnotetext{\textsuperscript{*}Equal contribution. \textsuperscript{\dag}Corresponding author: \href{mailto:sryoon@snu.ac.kr}{\nolinkurl{sryoon@snu.ac.kr}} \textsuperscript{\ddag}Now at Upstage AI.}
\renewcommand{\thefootnote}{\arabic{footnote}}
\fi

\input{sections/abstract}
\input{sections/introduction}
\input{sections/related_work}
\input{sections/method}
\input{sections/experiments}
\input{sections/discussion}
\input{sections/conclusion}

\bibliography{references}

\appendix
\input{sections/appendix}

\end{document}

%% file: macros.tex
\definecolor{cHl}{HTML}{E8F0F6}

\newcommand{\mbtr}{\mbox{MBT-R}}
\newcommand{\mbti}{\mbox{MBT-I}}
\newcommand{\mbta}{\mbox{MBT-A}}

\newcommand{\tgroup}[2]{\multicolumn{#1}{@{}l}{\textit{#2}}\\}

\newcommand{\best}[1]{\textbf{#1}}
\newcommand{\second}[1]{\underline{#1}}

\makeatletter
\newcommand{\setcaptype}[1]{\def\@captype{#1}}
\makeatother

%% file: tables/numbers.tex
\providecommand{\paperValue}[1]{\ifcsname paper-value:#1\endcsname\csname paper-value:#1\endcsname\else\PackageError{paper-data}{Unknown paper value: #1}{Rebuild data/results.json.}\fi}
\expandafter\def\csname paper-value:abstract-score/a\endcsname{85.4}
\expandafter\def\csname paper-value:abstract-score/u\endcsname{83.1}
\expandafter\def\csname paper-value:agent-cost/cost\endcsname{0.15}
\expandafter\def\csname paper-value:agent-cost/seconds\endcsname{872}
\expandafter\def\csname paper-value:agent-cost/turns\endcsname{52}
\expandafter\def\csname paper-value:agent-insertions/n\endcsname{10.8}
\expandafter\def\csname paper-value:agent-ks/ks\endcsname{0.161}
\expandafter\def\csname paper-value:agent-length-commits/commits\endcsname{15}
\expandafter\def\csname paper-value:agent-length-commits/ratio\endcsname{1.112}
\expandafter\def\csname paper-value:agent-repairs/commits\endcsname{397}
\expandafter\def\csname paper-value:agent-repairs/traces\endcsname{711}
\expandafter\def\csname paper-value:annotation-problem-sample/selected\endcsname{100}
\expandafter\def\csname paper-value:annotation-problem-sample/total\endcsname{153}
\expandafter\def\csname paper-value:appendix-answer-retries/n\endcsname{10}
\expandafter\def\csname paper-value:astro-score/score\endcsname{81.8}
\expandafter\def\csname paper-value:backtracking-insertions/a\endcsname{845}
\expandafter\def\csname paper-value:backtracking-insertions/i\endcsname{986}
\expandafter\def\csname paper-value:base-student-rates/cb\endcsname{30.5}
\expandafter\def\csname paper-value:base-student-rates/cu\endcsname{12.5}
\expandafter\def\csname paper-value:base-student-rates/sb\endcsname{35.0}
\expandafter\def\csname paper-value:base-student-rates/su\endcsname{11.5}
\expandafter\def\csname paper-value:batch-size/batch\endcsname{64}
\expandafter\def\csname paper-value:before-after-counts/correct\endcsname{100}
\expandafter\def\csname paper-value:before-after-counts/n\endcsname{200}
\expandafter\def\csname paper-value:before-after-counts/wrong\endcsname{100}
\expandafter\def\csname paper-value:benchmark-rollouts/math\endcsname{8}
\expandafter\def\csname paper-value:benchmark-rollouts/qa\endcsname{1}
\expandafter\def\csname paper-value:candidate-annotation-population/n\endcsname{40}
\expandafter\def\csname paper-value:clip-decay/clip\endcsname{1.0}
\expandafter\def\csname paper-value:clip-decay/decay\endcsname{0.0}
\expandafter\def\csname paper-value:clogit-pairs-appendix/n\endcsname{437}
\expandafter\def\csname paper-value:code-score/a\endcsname{50.3}
\expandafter\def\csname paper-value:code-score/u\endcsname{47.5}
\expandafter\def\csname paper-value:commits-appendix/empty\endcsname{1,216}
\expandafter\def\csname paper-value:commits-appendix/initial\endcsname{1,000}
\expandafter\def\csname paper-value:commits-appendix/insertions\endcsname{10,759}
\expandafter\def\csname paper-value:commits-appendix/repairs\endcsname{4,018}
\expandafter\def\csname paper-value:commits-appendix/total\endcsname{16,993}
\expandafter\def\csname paper-value:conclusion-cost/cost\endcsname{0.15}
\expandafter\def\csname paper-value:conclusion-cost/seconds\endcsname{872}
\expandafter\def\csname paper-value:conclusion-cost/turns\endcsname{52}
\expandafter\def\csname paper-value:content-repairs-appendix/n\endcsname{397}
\expandafter\def\csname paper-value:control-extra/extra\endcsname{62}
\expandafter\def\csname paper-value:control-training-population/n\endcsname{1,000}
\expandafter\def\csname paper-value:curve-rollouts/n\endcsname{8}
\expandafter\def\csname paper-value:decontamination-ngrams/n\endcsname{9}
\expandafter\def\csname paper-value:deletion-character-percentage/pct\endcsname{56.5}
\expandafter\def\csname paper-value:deletion-drops/d4\endcsname{16.4}
\expandafter\def\csname paper-value:deletion-drops/d8\endcsname{28.4}
\expandafter\def\csname paper-value:deletion-finished-count/finished\endcsname{998}
\expandafter\def\csname paper-value:deletion-finished-count/total\endcsname{1,000}
\expandafter\def\csname paper-value:deletion-scores/base\endcsname{77.8}
\expandafter\def\csname paper-value:deletion-scores/d4\endcsname{66.7}
\expandafter\def\csname paper-value:deletion-scores/d8\endcsname{54.7}
\expandafter\def\csname paper-value:deletion-scores/u\endcsname{83.1}
\expandafter\def\csname paper-value:deletion-start-count/n\endcsname{1,000}
\expandafter\def\csname paper-value:deletion-token-percentages/d4\endcsname{74.1}
\expandafter\def\csname paper-value:deletion-token-percentages/d8\endcsname{50.0}
\expandafter\def\csname paper-value:deletion-train-counts/delete\endcsname{993}
\expandafter\def\csname paper-value:deletion-train-counts/unedited\endcsname{995}
\expandafter\def\csname paper-value:diagnostic-caption/n\endcsname{200}
\expandafter\def\csname paper-value:diagnostic-codebook-size/desirable\endcsname{3}
\expandafter\def\csname paper-value:diagnostic-codebook-size/items\endcsname{16}
\expandafter\def\csname paper-value:diagnostic-codebook-size/types\endcsname{14}
\expandafter\def\csname paper-value:diagnostic-counts/correct\endcsname{105}
\expandafter\def\csname paper-value:diagnostic-counts/n\endcsname{200}
\expandafter\def\csname paper-value:diagnostic-counts/problems\endcsname{50}
\expandafter\def\csname paper-value:diagnostic-counts/training\endcsname{1,000}
\expandafter\def\csname paper-value:diagnostic-counts/wrong\endcsname{95}
\expandafter\def\csname paper-value:difficulty-boundaries/high\endcsname{8}
\expandafter\def\csname paper-value:difficulty-boundaries/low\endcsname{0}
\expandafter\def\csname paper-value:domain-gains/high\endcsname{0.6}
\expandafter\def\csname paper-value:domain-gains/low\endcsname{0.1}
\expandafter\def\csname paper-value:edit-measurements/edits\endcsname{776}
\expandafter\def\csname paper-value:editing-middle-share/mbt-a\endcsname{31.7}
\expandafter\def\csname paper-value:editing-middle-share/mbt-i\endcsname{20.8}
\expandafter\def\csname paper-value:epoch-crossover/early\endcsname{5}
\expandafter\def\csname paper-value:epoch-crossover/later\endcsname{10}
\expandafter\def\csname paper-value:epoch-table-checkpoints/first\endcsname{5}
\expandafter\def\csname paper-value:epoch-table-checkpoints/last\endcsname{15}
\expandafter\def\csname paper-value:epoch-table-checkpoints/second\endcsname{10}
\expandafter\def\csname paper-value:epochs-table/epochs\endcsname{15}
\expandafter\def\csname paper-value:extra-problems/n\endcsname{62}
\expandafter\def\csname paper-value:extra-tokens/pct\endcsname{7.1}
\expandafter\def\csname paper-value:failure-buckets/a\endcsname{0.0}
\expandafter\def\csname paper-value:failure-buckets/b\endcsname{3.8}
\expandafter\def\csname paper-value:failure-buckets/c\endcsname{19.4}
\expandafter\def\csname paper-value:failure-buckets/d\endcsname{40.4}
\expandafter\def\csname paper-value:failure-buckets/e\endcsname{38.0}
\expandafter\def\csname paper-value:failure-count-appendix/n\endcsname{224}
\expandafter\def\csname paper-value:figure-math-size/n\endcsname{153}
\expandafter\def\csname paper-value:generation-hyperparameters/context\endcsname{131,072}
\expandafter\def\csname paper-value:generation-hyperparameters/k\endcsname{-1}
\expandafter\def\csname paper-value:generation-hyperparameters/output\endcsname{122,880}
\expandafter\def\csname paper-value:generation-hyperparameters/p\endcsname{0.95}
\expandafter\def\csname paper-value:generation-hyperparameters/temp\endcsname{1.0}
\expandafter\def\csname paper-value:generation-retries/tries\endcsname{10}
\expandafter\def\csname paper-value:granite-score/a\endcsname{30.4}
\expandafter\def\csname paper-value:granite-score/u\endcsname{28.8}
\expandafter\def\csname paper-value:insertion-endpoints/first\endcsname{31.9}
\expandafter\def\csname paper-value:insertion-endpoints/last\endcsname{17.0}
\expandafter\def\csname paper-value:insertion-length/ratio\endcsname{1.048}
\expandafter\def\csname paper-value:insertion-middle/high\endcsname{6.0}
\expandafter\def\csname paper-value:insertion-middle/low\endcsname{2.5}
\expandafter\def\csname paper-value:intro-annotation-counts/desirable\endcsname{3}
\expandafter\def\csname paper-value:intro-annotation-counts/traces\endcsname{200}
\expandafter\def\csname paper-value:intro-annotation-counts/undesirable\endcsname{14}
\expandafter\def\csname paper-value:intro-control/control\endcsname{83.2}
\expandafter\def\csname paper-value:intro-diagnostic-gap/high\endcsname{19}
\expandafter\def\csname paper-value:intro-diagnostic-gap/low\endcsname{17}
\expandafter\def\csname paper-value:intro-main/a\endcsname{85.4}
\expandafter\def\csname paper-value:intro-main/u\endcsname{83.1}
\expandafter\def\csname paper-value:intro-prior-range/high\endcsname{84.2}
\expandafter\def\csname paper-value:intro-prior-range/low\endcsname{72.7}
\expandafter\def\csname paper-value:intro-rollouts/n\endcsname{8}
\expandafter\def\csname paper-value:intro-verification/pct\endcsname{95}
\expandafter\def\csname paper-value:judge-agreement-cohort/samples\endcsname{4,896}
\expandafter\def\csname paper-value:judge-agreement-cohort/ti-samples\endcsname{2,448}
\expandafter\def\csname paper-value:judge-agreement-floor/pct\endcsname{99.6}
\expandafter\def\csname paper-value:judge-agreement-range/high\endcsname{99.84}
\expandafter\def\csname paper-value:judge-agreement-range/low\endcsname{99.65}
\expandafter\def\csname paper-value:judge-problem-count/n\endcsname{153}
\expandafter\def\csname paper-value:last-tenth-controlled/a\endcsname{6.4}
\expandafter\def\csname paper-value:last-tenth-controlled/i\endcsname{2.9}
\expandafter\def\csname paper-value:learning-rate-warmup/rate\endcsname{1.0e-5}
\expandafter\def\csname paper-value:learning-rate-warmup/warmup\endcsname{40}
\expandafter\def\csname paper-value:livecodebench-release-size/n\endcsname{131}
\expandafter\def\csname paper-value:logical-check-association/odds\endcsname{2.60}
\expandafter\def\csname paper-value:logical-check-association/p\endcsname{0.022}
\expandafter\def\csname paper-value:main-ranking/a\endcsname{85.4}
\expandafter\def\csname paper-value:main-ranking/g\endcsname{84.2}
\expandafter\def\csname paper-value:main-ranking/i\endcsname{83.8}
\expandafter\def\csname paper-value:matched-pairs/all\endcsname{1,000}
\expandafter\def\csname paper-value:matched-pairs/used\endcsname{437}
\expandafter\def\csname paper-value:matched-token-ratios/a\endcsname{1.071}
\expandafter\def\csname paper-value:matched-token-ratios/control\endcsname{1.073}
\expandafter\def\csname paper-value:math-problem-count/n\endcsname{153}
\expandafter\def\csname paper-value:max-length-table/length\endcsname{131,072}
\expandafter\def\csname paper-value:max-student-length/length\endcsname{131,072}
\expandafter\def\csname paper-value:minicpm-score/a\endcsname{31.4}
\expandafter\def\csname paper-value:minicpm-score/u\endcsname{28.5}
\expandafter\def\csname paper-value:optimizer-hyperparameters/beta1\endcsname{0.9}
\expandafter\def\csname paper-value:optimizer-hyperparameters/beta2\endcsname{0.95}
\expandafter\def\csname paper-value:optimizer-hyperparameters/eps\endcsname{1e-8}
\expandafter\def\csname paper-value:opus-annotation-failures/empty\endcsname{8}
\expandafter\def\csname paper-value:opus-annotation-failures/malformed\endcsname{3}
\expandafter\def\csname paper-value:opus-annotation-failures/total\endcsname{40}
\expandafter\def\csname paper-value:opus-valid-annotations/n\endcsname{29}
\expandafter\def\csname paper-value:planning-schemes/correct\endcsname{17.1}
\expandafter\def\csname paper-value:planning-schemes/subgoal\endcsname{10.0}
\expandafter\def\csname paper-value:planning-schemes/wrong\endcsname{9.5}
\expandafter\def\csname paper-value:rare-behaviour-range/high\endcsname{11.0}
\expandafter\def\csname paper-value:rare-behaviour-range/low\endcsname{2.5}
\expandafter\def\csname paper-value:reference-annotation-population/n\endcsname{40}
\expandafter\def\csname paper-value:release-problem-count/n\endcsname{1,000}
\expandafter\def\csname paper-value:repeated-annotation-count/n\endcsname{40}
\expandafter\def\csname paper-value:rewrite-failed-cells/failed\endcsname{224}
\expandafter\def\csname paper-value:rewrite-failed-cells/total\endcsname{1,000}
\expandafter\def\csname paper-value:rewrite-failure-extremes/long\endcsname{40}
\expandafter\def\csname paper-value:rewrite-failure-extremes/short\endcsname{0}
\expandafter\def\csname paper-value:rewrite-failures/pct\endcsname{22.4}
\expandafter\def\csname paper-value:rewrite-long-ratio/ratio\endcsname{0.078}
\expandafter\def\csname paper-value:rewrite-no-headers/failed\endcsname{224}
\expandafter\def\csname paper-value:rewrite-no-headers/missing\endcsname{156}
\expandafter\def\csname paper-value:rewrite-score/r\endcsname{73.9}
\expandafter\def\csname paper-value:rewrite-score/u\endcsname{83.1}
\expandafter\def\csname paper-value:rewrite-shortening/n\endcsname{776}
\expandafter\def\csname paper-value:rewrite-shortening/pct\endcsname{28.1}
\expandafter\def\csname paper-value:rewrite-token-ratio/ratio\endcsname{0.331}
\expandafter\def\csname paper-value:search-problem-count/n\endcsname{153}
\expandafter\def\csname paper-value:seed-comparison-epoch/n\endcsname{15}
\expandafter\def\csname paper-value:seed-ranges-appendix/amax\endcsname{85.4}
\expandafter\def\csname paper-value:seed-ranges-appendix/amin\endcsname{85.0}
\expandafter\def\csname paper-value:seed-ranges-appendix/umax\endcsname{83.7}
\expandafter\def\csname paper-value:seed-ranges-appendix/umin\endcsname{82.8}
\expandafter\def\csname paper-value:seed-ranges/amax\endcsname{85.4}
\expandafter\def\csname paper-value:seed-ranges/amin\endcsname{85.0}
\expandafter\def\csname paper-value:seed-ranges/umax\endcsname{83.7}
\expandafter\def\csname paper-value:seed-ranges/umin\endcsname{82.8}
\expandafter\def\csname paper-value:selected-training-recipe/epochs\endcsname{15}
\expandafter\def\csname paper-value:selected-training-recipe/lr\endcsname{10^{-5}}
\expandafter\def\csname paper-value:self-correction-counts/a\endcsname{704}
\expandafter\def\csname paper-value:self-correction-counts/i\endcsname{120}
\expandafter\def\csname paper-value:serving-software/flashinfer\endcsname{0.6.18.post1}
\expandafter\def\csname paper-value:serving-software/torch\endcsname{2.13.0+cu130}
\expandafter\def\csname paper-value:serving-software/vllm\endcsname{0.30.0}
\expandafter\def\csname paper-value:setup-answer-retries/n\endcsname{10}
\expandafter\def\csname paper-value:setup-checkpoint-repeated/epoch\endcsname{15}
\expandafter\def\csname paper-value:setup-epoch/epoch\endcsname{15}
\expandafter\def\csname paper-value:student-behaviour-caption/n\endcsname{200}
\expandafter\def\csname paper-value:student-behaviour-rates/ba\endcsname{42.0}
\expandafter\def\csname paper-value:student-behaviour-rates/bu\endcsname{9.0}
\expandafter\def\csname paper-value:student-behaviour-rates/ca\endcsname{96.5}
\expandafter\def\csname paper-value:student-behaviour-rates/cu\endcsname{12.5}
\expandafter\def\csname paper-value:student-behaviour-rates/sa\endcsname{87.5}
\expandafter\def\csname paper-value:student-behaviour-rates/su\endcsname{11.5}
\expandafter\def\csname paper-value:student-counts/problems\endcsname{100}
\expandafter\def\csname paper-value:student-counts/traces\endcsname{200}
\expandafter\def\csname paper-value:student-transfer-caption/traces\endcsname{200}
\expandafter\def\csname paper-value:subsample-count/n\endcsname{510}
\expandafter\def\csname paper-value:subsample-gain/gain\endcsname{16.9}
\expandafter\def\csname paper-value:subsample-score/score\endcsname{83.6}
\expandafter\def\csname paper-value:table-annotator/opus\endcsname{29}
\expandafter\def\csname paper-value:table-annotator/reference\endcsname{40}
\expandafter\def\csname paper-value:table-annotator/repeat\endcsname{40}
\expandafter\def\csname paper-value:table-baselines/bridge-cells\endcsname{1,000}
\expandafter\def\csname paper-value:table-baselines/bridged-traces\endcsname{529}
\expandafter\def\csname paper-value:table-baselines/bridges\endcsname{1,209}
\expandafter\def\csname paper-value:table-baselines/bridges-per-trace\endcsname{2.3}
\expandafter\def\csname paper-value:table-baselines/gandhi-changed\endcsname{96}
\expandafter\def\csname paper-value:table-baselines/gandhi-changed-pct\endcsname{9.6}
\expandafter\def\csname paper-value:table-baselines/gandhi-traces\endcsname{998}
\expandafter\def\csname paper-value:table-diagnostic/correct\endcsname{105}
\expandafter\def\csname paper-value:table-diagnostic/traces\endcsname{200}
\expandafter\def\csname paper-value:table-diagnostic/wrong\endcsname{95}
\expandafter\def\csname paper-value:table-editing/rewrite-edited\endcsname{776}
\expandafter\def\csname paper-value:table-math/problems\endcsname{153}
\expandafter\def\csname paper-value:table-paired/correct\endcsname{100}
\expandafter\def\csname paper-value:table-paired/traces\endcsname{200}
\expandafter\def\csname paper-value:table-paired/wrong\endcsname{100}
\expandafter\def\csname paper-value:table-repairs/traces\endcsname{711}
\expandafter\def\csname paper-value:table-traces/n\endcsname{1,000}
\expandafter\def\csname paper-value:table-train-data/deletion-tokens\endcsname{22,688,991}
\expandafter\def\csname paper-value:table-train-data/deletion-traces\endcsname{993}
\expandafter\def\csname paper-value:table-train-data/max-length\endcsname{131,072}
\expandafter\def\csname paper-value:target-codebook-cardinality/n\endcsname{7}
\expandafter\def\csname paper-value:target-reliability/smax\endcsname{0.87}
\expandafter\def\csname paper-value:target-reliability/smin\endcsname{0.49}
\expandafter\def\csname paper-value:target-reliability/tmax\endcsname{1.00}
\expandafter\def\csname paper-value:target-reliability/tmin\endcsname{0.72}
\expandafter\def\csname paper-value:teacher-accuracy/pct\endcsname{58.5}
\expandafter\def\csname paper-value:teacher-behaviour-rates/ba\endcsname{55.0}
\expandafter\def\csname paper-value:teacher-behaviour-rates/bb\endcsname{2.5}
\expandafter\def\csname paper-value:teacher-behaviour-rates/ca\endcsname{100}
\expandafter\def\csname paper-value:teacher-behaviour-rates/cb\endcsname{11.0}
\expandafter\def\csname paper-value:teacher-behaviour-rates/sa\endcsname{99.5}
\expandafter\def\csname paper-value:teacher-behaviour-rates/sb\endcsname{10.0}
\expandafter\def\csname paper-value:teacher-common-problems/n\endcsname{50}
\expandafter\def\csname paper-value:teacher-figure-counts/problems\endcsname{50}
\expandafter\def\csname paper-value:teacher-figure-counts/traces\endcsname{200}
\expandafter\def\csname paper-value:teacher-length/max\endcsname{150,834}
\expandafter\def\csname paper-value:teacher-length/mean\endcsname{22,965}
\expandafter\def\csname paper-value:teacher-pair-counts/correct\endcsname{100}
\expandafter\def\csname paper-value:teacher-pair-counts/n\endcsname{200}
\expandafter\def\csname paper-value:teacher-pair-counts/wrong\endcsname{100}
\expandafter\def\csname paper-value:ti-score/base\endcsname{77.8}
\expandafter\def\csname paper-value:ti-score/ti\endcsname{72.7}
\expandafter\def\csname paper-value:token-control-score/control\endcsname{83.2}
\expandafter\def\csname paper-value:token-control-score/u\endcsname{83.1}
\expandafter\def\csname paper-value:training-bin-counts/high\endcsname{112}
\expandafter\def\csname paper-value:training-bin-counts/low\endcsname{111}
\expandafter\def\csname paper-value:training-bin-counts/total\endcsname{1,000}
\expandafter\def\csname paper-value:training-pool-size/n\endcsname{10,000}
\expandafter\def\csname paper-value:training-software/accelerate\endcsname{1.15.0}
\expandafter\def\csname paper-value:training-software/liger\endcsname{0.8.3}
\expandafter\def\csname paper-value:training-software/python\endcsname{3.12.13}
\expandafter\def\csname paper-value:training-software/torch\endcsname{2.14.0}
\expandafter\def\csname paper-value:training-software/transformers\endcsname{5.17.0}
\expandafter\def\csname paper-value:training-software/triton\endcsname{3.8.0}
\expandafter\def\csname paper-value:training-software/trl\endcsname{1.14.0}
\expandafter\def\csname paper-value:transfer-caption/teacher\endcsname{200}
\expandafter\def\csname paper-value:verification-association/n\endcsname{14}
\expandafter\def\csname paper-value:verification-association/p\endcsname{0.15}
\expandafter\def\csname paper-value:verification-discordant/n\endcsname{14}

%% file: sections/abstract.tex
\begin{abstract}
Distilling long reasoning traces from a teacher model into a smaller student also passes on the teacher's reasoning habits.
Prior work and our annotation of teacher traces show that current reasoning models often start without a plan, verify superficially, abandon promising solution paths, and take unchecked shortcuts.
We propose \textbf{M}etacognitive \textbf{B}ehavioral \textbf{T}uning (MBT), which inserts seven metacognitive behaviors into teacher traces at the points where the reasoning lacks them and then trains the student on the edited traces.
Because teacher traces are long, how the edits are made matters. Rewriting a whole trace at once summarizes long traces, and choosing all insertion points at once crowds them at the start and end.
Our agentic method, \mbta{}, edits the trace file line by line and records each edit as a git commit, which avoids summarizing long traces and spreads the insertions.
On competition math, a Qwen3.5-4B-Base student trained on \mbta{} traces scores \paperValue{abstract-score/a} against \paperValue{abstract-score/u} for SFT on the unedited traces and also outperforms a token-matched control and five prior methods. \mbta{} improves two other students as well, and the inserted behaviors clearly transfer to the students' reasoning, with rare ones several times more frequent.
\end{abstract}

%% file: sections/introduction.tex
\section{Introduction}
\label{sec:intro}

Distilling long reasoning traces, the step-by-step reasoning a model writes before its final answer, from a strong teacher model into a smaller student model through supervised fine-tuning (SFT) has become a standard way to improve math and code reasoning \citep{deepseek2025r1,muennighoff2025s1,ye2025limo,guha2025openthoughts}.
The student learns not only the final answers but the whole reasoning process, so the reasoning patterns of the teacher's traces, desirable or not, carry over to the student \citep{li2026patterns}.
This raises a natural question. Instead of distilling teacher traces as they are, can we first improve the reasoning behaviors they demonstrate?

Prior work has documented recurring weaknesses in the traces of current reasoning models (Appendix~\ref{app:prior}).
DeepSeek-R1 plans as it executes rather than planning first \citep{marjanovic2025thoughtology}, although planning sentences are among those that most influence the final answer \citep{bogdan2025thought}.
In failed DeepSeek-R1 traces, about three quarters of reflections are surface checks that miss the error, while reflections that check the logic of a deductive step are strongly associated with correct traces \citep{chen2026anatomy}.
Many incorrect traces contain a correct idea that is abandoned \citep{wang2025underthinking}, and models often miss errors in their own output that they correct in the user's input \citep{tsui2025selfcorrection}.

These studies also suggest that \emph{where} and \emph{how} a behavior appears matters more than \emph{how often}. Correct and incorrect traces do not differ on average in how often they reflect or backtrack, but logical checks inside deductive steps mark correct traces \citep{chen2026anatomy}, and the structure of a trace predicts correctness better than its length \citep{jiang2025structure}.
A reflection marker such as ``Wait'' is not itself a reflection. Suppressing such markers at inference shortens traces at little cost in accuracy \citep{wang2025wait}.

We find related weaknesses in our teacher, DeepSeek-V4.1-Flash \citep{deepseek2026v41flash}, on math problems from Nemotron-Math-v2 \citep{du2025nemotronmath}.
An LLM annotator (gpt-6.1-sol, \citealp{openai2026sol}), which sees the problem and the reasoning but not whether the answer is correct, annotates \paperValue{intro-annotation-counts/traces} teacher traces for \paperValue{intro-annotation-counts/undesirable} undesirable and \paperValue{intro-annotation-counts/desirable} desirable behavior types (Appendix~\ref{app:teacher}).
Behaviors such as guessing the answer the problem setter expects, overthinking after an answer is reached, and asserting a recalled answer without checking it are \paperValue{intro-diagnostic-gap/low} to \paperValue{intro-diagnostic-gap/high} percentage points more frequent in incorrect traces (Figure~\ref{fig:teacher-behaviors}), although harder problems raise both these behaviors and errors.
In contrast, about \paperValue{intro-verification/pct}\% of both correct and incorrect traces contain at least one substantive verification, so the mere presence of verification does not separate them (an example trace is in Figure~\ref{fig:teacher-trace}).

\input{figures/teacher_behaviors}
\input{figures/overview}

We propose \textbf{M}etacognitive \textbf{B}ehavioral \textbf{T}uning (MBT), which inserts metacognitive behaviors that address these weaknesses into teacher traces and trains the student on the edited traces (Figure~\ref{fig:overview}).\footnote{\url{\repourl}}
We define seven \emph{target behaviors}, namely subgoal setting, logical check, substantive verification, targeted backtracking, resuming abandoned paths, self-correction, and sufficiency judgment.
Instead of requesting each behavior a fixed number of times, MBT specifies where each behavior belongs, for example a logical check right after a deductive step whose premise could fail.
Every inserted passage must state what it checked and what it found.
Unlike correctness filtering, MBT keeps incorrect traces, and its editor never sees the correct answer.

Making such edits automatically is difficult because teacher traces are long (reasoning sections average \paperValue{teacher-length/mean} tokens, up to \paperValue{teacher-length/max}) and the right editing position depends on the surrounding reasoning.
We compare three strategies that share the editor model and instructions, namely whole-trace rewriting (\mbtr{}), marker-based insertion (\mbti{}), and agentic editing (\mbta{}).
Rewriting the whole trace at once fails on \paperValue{rewrite-failures/pct}\% of the traces and summarizes long ones, and choosing all insertion points at once crowds them at the start and end.
\mbta{} addresses both problems by treating a trace as a file in a git repository, where an agent inserts passages after lines and commits each edit.

On five competition-math benchmarks, a Qwen3.5-4B-Base \citep{qwen2026qwen35} student trained on \mbta{} traces scores \paperValue{intro-main/a} (avg@\paperValue{intro-rollouts/n}), compared with \paperValue{intro-main/u} for SFT on the unedited traces (\emph{Unedited SFT}), \paperValue{intro-control/control} for a token-matched control, and \paperValue{intro-prior-range/low} to \paperValue{intro-prior-range/high} for five prior behavior-inducing methods. \mbta{} also improves over Unedited SFT for two other students.
The target behaviors that are rare in the original traces become frequent in the edited traces and in the students' reasoning.

Our contributions are as follows:
\begin{itemize}[nosep,leftmargin=*]
\item Combining prior findings with a comparison of correct and incorrect teacher traces, we identify undesirable behaviors and select seven metacognitive behaviors to address them.
\item We compare three ways of inserting these behaviors into long traces and introduce \mbta{}, which treats a trace as a code file under version control and edits it locally, line by line.
\item We show that \mbta{} traces yield better students than unedited traces, a token-matched control, and five prior methods, and that the inserted behaviors transfer to the students' reasoning.
\end{itemize}

%% file: figures/teacher_behaviors.tex
\begin{figure}[t]
\centering
\includegraphics[width=\columnwidth]{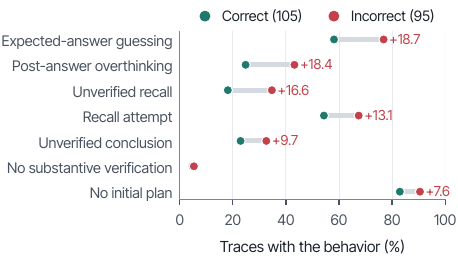}
\caption{Undesirable behaviors in \paperValue{teacher-figure-counts/traces} diagnostic teacher traces (\paperValue{teacher-figure-counts/problems} problems, four each). Shortcuts and overthinking are more frequent in incorrect traces, while almost no trace lacks substantive verification.}
\label{fig:teacher-behaviors}
\end{figure}

%% file: figures/overview.tex
\begin{figure*}[t]
\centering
\includegraphics[width=\textwidth]{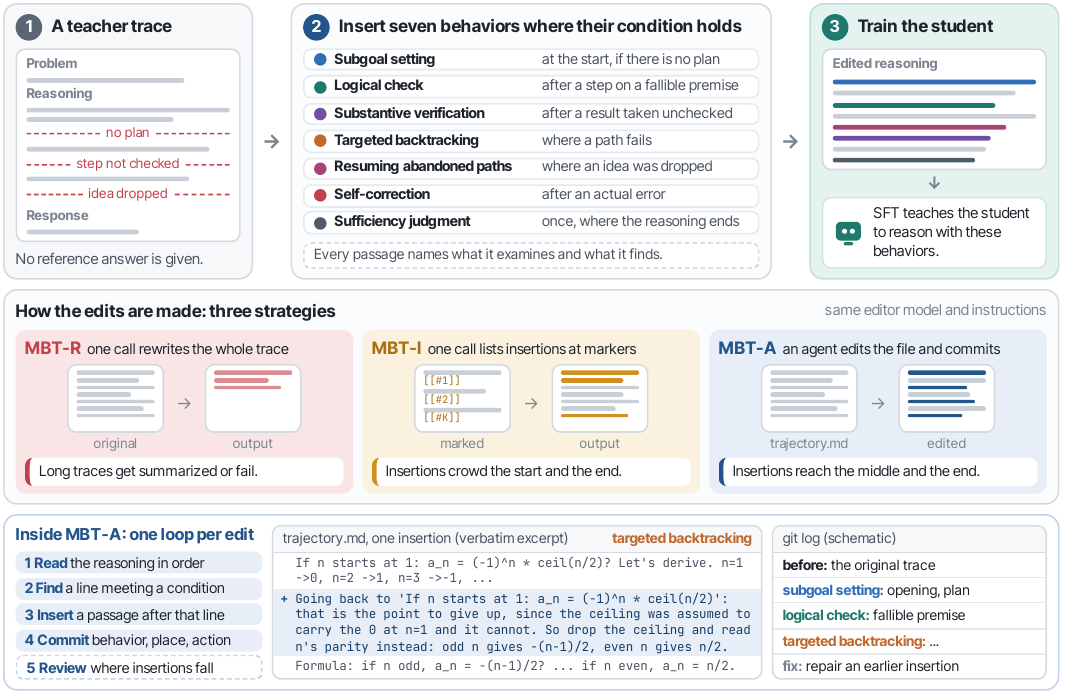}
\caption{MBT inserts target behaviors into teacher traces and trains the student on the edited traces. The three editing strategies share the editor model and instructions and differ in how the edits are made (the bottom row shows the \mbta{} loop and one real insertion).}
\label{fig:overview}
\end{figure*}

%% file: sections/related_work.tex
\section{Related Work}
\label{sec:related}

\paragraph{Distillation of long reasoning traces.}
SFT on the traces of strong reasoning models yields large gains even from about a thousand problems \citep{muennighoff2025s1,ye2025limo}. OpenThoughts \citep{guha2025openthoughts} finds that filtering by LLM verification, majority consensus, or length does not beat using all answers and that higher-scoring teachers are not necessarily better.
\citet{li2026patterns} trace such teacher effects to reasoning patterns. Students trained on DeepSeek-R1 traces, which branch frequently, score lower than students trained on gpt-oss traces despite a lower training loss, and removing frequently branching traces helps while removing the longest traces lowers accuracy.
These studies choose which traces to keep, while we revise the content of the traces.

\paragraph{Metacognition and cognitive behaviors.}
Metacognition is the monitoring and regulation of one's own cognition \citep{flavell1979,nelson1990,schraw1995}.
In mathematical problem solving \citep{polya1945}, experts check their progress and change course while novices pursue one direction unchecked, so novices fail for lack of monitoring despite having the needed knowledge \citep{schoenfeld1985}. Similar episodes appear in the traces of DeepSeek-R1 \citep{li2025schoenfeld}.
\citet{gandhi2025cognitive} attribute the gap between two models under reinforcement learning to four cognitive behaviors (verification, backtracking, subgoal setting, and backward chaining), and priming the weaker model with traces that show them closes the gap on Countdown even with traces that reach wrong answers.
\citet{kargupta2025cognitive} show that models often do not use the behaviors a problem needs, although they use them elsewhere, and prompting for them at inference helps only capable models. We instead teach these behaviors through training data.

\paragraph{Methods for inducing reasoning behaviors.}
ASTRO \citep{kim2025astro} linearizes Monte Carlo tree search into one solution and inserts fixed reflection phrases where the search backtracks.
CoT-Bridge \citep{xu2025mindgap} fills in missing intermediate steps, and RefAug \citep{zhang2024refaug} appends a reflection after the answer for training only.
\citet{gandhi2025cognitive} generate new solutions with a prompt that asks for the four behaviors, and Thinking Intervention (TI, \citealp{wu2025thinking}) inserts a first-person instruction at the start of reasoning without training.
MBT instead inserts substantive passages for seven behaviors where existing traces lack them, choosing each position from the surrounding reasoning. Section~\ref{sec:experiments} compares all five.

%% file: sections/method.tex
\section{Metacognitive Behavioral Tuning}
\label{sec:method}

\subsection{Problem Setup}
\label{sec:problem-setup}

A training example is a triple $(x, r, a)$, where $x$ is a problem, $r$ is the teacher's \emph{reasoning section} (the text inside the thinking block), and $a$ is the teacher's \emph{response section} (the text after it, which contains the final answer).
The original data are $\mathcal{D} = \{(x_i, r_i, a_i)\}_{i=1}^{N}$, and $\mathcal{B} = \{b_1, \dots, b_\paperValue{target-codebook-cardinality/n}\}$ is the set of target behaviors (Figure~\ref{fig:overview}).
An editing strategy $m \in \{\mathrm{R}, \mathrm{I}, \mathrm{A}\}$ is a procedure $E_m$ that calls an \emph{editor model} and returns edited reasoning and response sections, $(r'_i, a'_i) = E_m(x_i, r_i, a_i; \mathcal{B})$.

The editor model receives only the problem, the two sections, and the instructions, never the reference answer, and may not look the problem up. An editor that knew the answer could place corrections exactly where the trace disagrees with it and so leak the answer into the training data.
The edited data are $\mathcal{D}'_m = \{(x_i, r'_i, a'_i)\}_{i \in \mathcal{I}_m}$, where $\mathcal{I}_m \subseteq \{1, \dots, N\}$ contains the successfully edited examples that fit the maximum training length.

SFT on the unedited traces, which we call \emph{Unedited SFT}, is the case $(r'_i, a'_i) = (r_i, a_i)$.
The student $p_\theta$ with parameters $\theta$ is trained on the token sequence $y_i = (r'_i, a'_i) = (y_{i,1}, \dots, y_{i,T_i})$ with the loss on these assistant tokens only,
\begin{equation}
\mathcal{L}(\theta) = -\frac{1}{T} \sum_{i \in \mathcal{I}_m} \sum_{t=1}^{T_i} \log p_\theta\bigl(y_{i,t} \mid x_i, y_{i,<t}\bigr),
\label{eq:loss}
\end{equation}
where $T_i$ is the number of assistant tokens of example $i$, $T = \sum_{i \in \mathcal{I}_m} T_i$, and $y_{i,<t} = (y_{i,1}, \dots, y_{i,t-1})$. In training, the average is taken in the same way within each minibatch.
All three strategies use the teacher itself, DeepSeek-V4.1-Flash, as the editor model and share the same instructions for the behaviors. They differ only in how the editor works and what it outputs.

\subsection{Target Behaviors}
\label{sec:target}

\input{tables/target_compact}

Each target behavior addresses undesirable behaviors found by prior work and our annotation (Table~\ref{tab:target-compact}, Appendices~\ref{app:prior} and~\ref{app:teacher}).
A trace that starts without a plan receives subgoal setting. A conclusion drawn from an unchecked calculation receives a substantive verification, for example substituting the answer back into the problem. A trace that guesses the expected answer or quotes a remembered one receives both a substantive verification and a logical check that asks whether the answer actually follows from the preceding steps, and a trace that keeps exploring after it has an answer receives a sufficiency judgment that collects the grounds and settles the answer.

Subgoal setting, targeted backtracking, and sufficiency judgment occur in only \paperValue{rare-behaviour-range/low}\% to \paperValue{rare-behaviour-range/high}\% of the original traces and so have the most room to grow, whereas common behaviors such as substantive verification are added only where they are missing, for example after an unchecked result.

The editing instructions follow three principles (Appendix~\ref{app:instructions}).
(1) They never ask for a number of insertions. Each behavior comes with a description of where it belongs, with examples.
For instance, subgoal setting goes right after the opening, and only into traces whose opening says what to find but not what must be established first. Resuming abandoned paths goes where the trace raises a possibility, derives nothing from it, and moves on.
If no position fits the description, the behavior is not inserted.
(2) An inserted passage must name what it examined and state what it found.
``Wait, let me check.'' alone is not allowed, whereas the example in the instructions, ``Substituting back into the original condition $x^2 - 8m^2 = -7$ gives $81 - 88 = -7$, so the pair satisfies it,'' names the object of the check and its result.
(3) A behavior is not inserted where the trace already shows it, so that edits do not repeat what the trace already does.

\subsection{Three Editing Strategies}
\label{sec:strategies}

Figure~\ref{fig:overview} compares the three strategies. Their output formats are in Appendix~\ref{app:instructions}.

\paragraph{Whole-trace rewriting (\mbtr{}).}
$E_{\mathrm{R}}$ receives the whole trace with the instructions and, in a single call, outputs the entire edited reasoning and response sections. This is the most direct approach, but the editor must reproduce tens of thousands of tokens of original text in one output while also deciding where to insert.
An acceptance function $\mathrm{acc}(o) \in \{0, 1\}$ returns 1 if an output $o$ has non-empty reasoning and response sections in the required format and no line that replaces original text with a placeholder such as ``(rest omitted).''
The editor is called until an output is accepted, at most 10 times. With $o_j$ the output of the $j$-th call, the edit is the first accepted output $o_{j^\star}$ with
\begin{equation}
j^\star = \min\{\, j \le 10 : \mathrm{acc}(o_j) = 1 \,\},
\end{equation}
and the edit fails if no such $j$ exists. Calls also stop after five consecutive outputs hit the length limit.

\paragraph{Marker-based insertion (\mbti{}).}
The reasoning section is split at blank lines into paragraphs $(c_1, \dots, c_K)$, and a numbered marker \texttt{[[\#k]]} is appended to each paragraph $c_k$. A note before the trace states that there are $K$ markers and that \texttt{[[\#K]]} ends the reasoning section.
In a single call (retried if the output is malformed), the editor returns only an insertion list $S = \{(k_j, \tilde{b}_j, s_j)\}_{j=1}^{|S|}$, where $k_j \in \{1, \dots, K\}$ is a marker, $\tilde{b}_j \in \mathcal{B}$ a behavior, and $s_j$ the passage.
Each $s_j$ is inserted by rule after $c_{k_j}$ (in list order when several share a marker), and the response section is kept ($a' = a$).
The original text is preserved by construction, but the editor must choose every position in a single pass over a long trace. Section~\ref{sec:edit-compare} shows that it then favors the beginning and the end.

\paragraph{Agentic editing (\mbta{}).}
\mbta{} turns trace editing into file editing in a code repository, a task that coding agents are built for, since an agent can read a long file in pieces, change a few lines at a time without reproducing the rest, and come back to an earlier place.
The trace is a single file, \texttt{trajectory.md}, in a git repository whose first commit holds the unedited trace.
An agent runs in the Claude Code harness \citep{anthropic2025claudecode} in an isolated sandbox without web search or web fetch tools. It uses the same editor model, so the same model writes the passages in all strategies.

The agent reads the reasoning section, inserts a passage after the line where a behavior belongs, and records each edit as one commit whose subject names the behavior, the position, and what was done. The instructions forbid deleting an incorrect step and ask for its correction after it.
The agent can repair passages it inserted earlier, and at the end it checks where its insertions fall and rereads parts it may have skipped. This review asks the agent to look again, not to spread its insertions evenly.
If the edited reasoning reaches a different conclusion, the agent revises the response section, and otherwise $a' = a$.
The commit history records each passage's behavior, position, and what was done (Appendix~\ref{app:instructions}).

%% file: tables/target_compact.tex
\begin{table}[t]
\centering
\footnotesize
\setlength{\tabcolsep}{3pt}
\renewcommand{\arraystretch}{1.05}
\begin{tabularx}{\linewidth}{@{}>{\raggedright\arraybackslash}p{1.95cm}>{\raggedright\arraybackslash}Xr@{}}
\toprule
Target behavior & Undesirable behaviors it addresses & \makecell[r]{Orig.\\(\%)} \\
\midrule
Subgoal setting & no initial plan & 10.0 \\
Logical check & missing logical check, expected-answer guessing, unverified recall & 77.0 \\
Substantive verification & superficial verification, no substantive verification, unverified conclusion, expected-answer guessing, unverified recall, recall attempt & 96.0 \\
Targeted backtracking & shallow backtracking & 2.5 \\
Resuming abandoned paths & abandoning promising paths & 76.5 \\
Self-correction & failure to self-correct, unverified conclusion & 64.0 \\
Sufficiency judgment & post-answer overthinking, unchecked final answer, expected-answer guessing & 11.0 \\
\bottomrule
\end{tabularx}
\caption{Target behaviors, the undesirable behaviors each addresses, and the share of \paperValue{table-paired/traces} original teacher traces that already contain it. Where each is inserted is shown in Figure~\ref{fig:overview}, and definitions are in Table~\ref{tab:target-behaviors}.}
\label{tab:target-compact}
\end{table}

%% file: sections/experiments.tex
\section{Experiments}
\label{sec:experiments}

\subsection{Experimental Setup}
\label{sec:setup}

\paragraph{Data.}
From Nemotron-Math-v2 we take a pool of \paperValue{training-pool-size/n} problems after deduplication and decontamination, divide it into nine difficulty bins by how many of eight gpt-oss-120b \citep{openai2025gptoss} attempts are correct (\paperValue{difficulty-boundaries/low} to \paperValue{difficulty-boundaries/high}), and sample \paperValue{training-bin-counts/low} or \paperValue{training-bin-counts/high} problems per bin, \paperValue{training-bin-counts/total} in total.
Decontamination removes problems that match a benchmark problem after normalization or share five or more word \paperValue{decontamination-ngrams/n}-grams with one, a looser form of the Nemotron-Cascade check \citep{wang2025nemotroncascade}, which removes a problem sharing one 9-gram (details in Appendix~\ref{app:setup}).
The teacher solves each problem once (\paperValue{teacher-accuracy/pct}\% correct), and we keep incorrect traces.

\input{figures/editing}

\paragraph{Students and training.}
The main student is Qwen3.5-4B-Base.
To test whether the results depend on this model, we also train MiniCPM5-2B-Base \citep{openbmb2026minicpm5} and Granite 4.1 3B Base \citep{ibm2026granite}.
All conditions differ only in training data and share settings chosen for Unedited SFT. We report the checkpoint after \paperValue{setup-epoch/epoch} epochs (Appendix~\ref{app:setup}).

\paragraph{Evaluation.}
For math, each student solves the \paperValue{math-problem-count/n} problems of AIME 2025, AIME 2026, HMMT February 2025, HMMT February 2026, and HMMT November 2025 released by MathArena \citep{balunovic2025matharena}, eight times each.
A sample from which no answer can be extracted is regenerated up to \paperValue{setup-answer-retries/n} times and scored 0 if it still has none.
An LLM judge, gpt-5.6-luna \citep{openai2026luna}, grades the answers. Its verdicts agree with the rule-based checker math\_verify \citep{kydlicek2025mathverify} on at least \paperValue{judge-agreement-floor/pct}\% of samples (Appendix~\ref{app:setup}).
Since the training data contain only math, we also evaluate on untrained domains, namely science (GPQA Diamond, \citealp{rein2023gpqa}), code (LiveCodeBench v6, \citealp{jain2024livecodebench}), and multi-hop question answering over several documents (MuSiQue, HotpotQA, and 2WikiMultihopQA, \citealp{trivedi2022musique,yang2018hotpotqa,ho2020constructing}).

For a problem set $\mathcal{Q}$ and $n$ samples $\hat{y}_{q,1}, \dots, \hat{y}_{q,n}$ per problem $q$, the grader (LLM judge, or tests for LiveCodeBench) returns $G(\hat{y}, y^{\star}_q) \in \{0,1\}$ against the reference $y^{\star}_q$, and we report
\begin{equation}
\mathrm{avg@}n = \frac{1}{|\mathcal{Q}|} \sum_{q \in \mathcal{Q}} \frac{1}{n} \sum_{j=1}^{n} G(\hat{y}_{q,j}, y^{\star}_q),
\end{equation}
in percent, with $n = \paperValue{benchmark-rollouts/math}$ for math, GPQA, and LiveCodeBench and $n = \paperValue{benchmark-rollouts/qa}$ for multi-hop QA.

\paragraph{Compared conditions.}
Besides Unedited SFT and the three MBT strategies, we train a \emph{token-matched control}, which adds to the original data the unedited traces of \paperValue{extra-problems/n} more problems sampled in the same way, so that it has as many training tokens as \mbta{} (\paperValue{matched-token-ratios/control} and \paperValue{matched-token-ratios/a} times the original).
We reimplement the five methods of Section~\ref{sec:related}, staying as close to the original papers as possible. ASTRO, CoT-Bridge, RefAug, and Gandhi et al.\ build training data from the same problems (and, except Gandhi et al., the same traces), and TI is applied to the Base model at inference.

Wherever a method needs a model call, it uses the original paper's prompt with our editor model and settings, so that differences in editor capability do not enter the comparison (Appendix~\ref{app:baselines}).

\subsection{Comparison of the Editing Strategies}
\label{sec:edit-compare}

Before training any student, we ask where each strategy places its insertions, whether it keeps the length of the trace, and whether its edits succeed. Figure~\ref{fig:editing} answers each in one panel.
Panel (a) shows where insertions fall. An insertion's position is $\pi = \ell / |r|_{\mathrm{chr}} \in [0,1]$, where $\ell$ counts the characters of the original reasoning before it and $|r|_{\mathrm{chr}}$ all of them, and the panel gives the share of insertions in each tenth of $[0,1]$, which would be 10\% everywhere under uniform placement.
Panel (b) shows the mean length ratio $\rho_i = |r'_i|_{\mathrm{tok}} / |r_i|_{\mathrm{tok}}$ (Qwen3.5 tokens) by original length, where a ratio below 1 means the edit shortens the trace.
Panel (c) shows how often \mbtr{} fails to produce an acceptable edit.
We summarize how evenly a strategy spreads its insertions with the Kolmogorov-Smirnov statistic $D$ against uniform placement \citep{massey1951} (lower is more even, Figure~\ref{fig:editing} legend). An even spread is not a goal in itself, but strong concentration points to parts the editor does not consider (definition and all statistics in Appendix~\ref{app:editing}).

\input{tables/main}

\paragraph{\mbtr{} fails on and summarizes long traces.}
\mbtr{} produces no acceptable output within at most 10 calls for \paperValue{rewrite-failed-cells/failed} of the \paperValue{rewrite-failed-cells/total} traces, with failures rising from \paperValue{rewrite-failure-extremes/short}\% below 4k tokens to about \paperValue{rewrite-failure-extremes/long}\% above 16k (Figure~\ref{fig:editing}c).
Most failures (\paperValue{rewrite-no-headers/missing} of \paperValue{rewrite-no-headers/failed}) lack section headers because the editor copies the trace into its thinking and then goes on solving the problem instead of returning the edit (Appendix~\ref{app:editing}).
Of the \paperValue{rewrite-shortening/n} successful edits, \paperValue{rewrite-shortening/pct}\% are shorter than half the original, and traces of 32k tokens or more shrink to a median ratio of \paperValue{rewrite-long-ratio/ratio} (Table~\ref{tab:editing-full}).
The outputs behave like summaries, leaving \mbtr{} \paperValue{rewrite-token-ratio/ratio} of the original tokens.

\paragraph{\mbti{} keeps the length but crowds the ends.}
\mbti{} keeps the length (median ratio \paperValue{insertion-length/ratio}), but its insertions go to the start and end, with \paperValue{insertion-endpoints/first}\% in the first tenth and \paperValue{insertion-endpoints/last}\% in the last but only \paperValue{insertion-middle/low}\% to \paperValue{insertion-middle/high}\% in each tenth between 40\% and 90\% (Figure~\ref{fig:editing}a).
This resembles the lost-in-the-middle effect, where language models use information in the middle of a long input less well than at its ends \citep{liu2024lost}.
The last tenth is large because the sufficiency judgment belongs at the end. Excluding sufficiency judgment and subgoal setting, only \paperValue{last-tenth-controlled/i}\% of \mbti{}'s insertions fall in the last tenth against \paperValue{last-tenth-controlled/a}\% of \mbta{}'s.

\paragraph{\mbta{} spreads its insertions most evenly.}
\mbta{} (median length ratio \paperValue{agent-length-commits/ratio}, a median of \paperValue{agent-length-commits/commits} commits per trace) makes the most insertions (\paperValue{agent-insertions/n} per trace on average) and has the smallest KS statistic (\paperValue{agent-ks/ks}), and for five of the seven behaviors its median insertion position lies later in the trace than \mbti{}'s (Appendix~\ref{app:editing}).
The relative difference is largest for self-correction, which requires finding an actual error first (\paperValue{self-correction-counts/a} insertions by \mbta{} against \paperValue{self-correction-counts/i} by \mbti{}).
The agent also repairs its own earlier insertions in \paperValue{agent-repairs/traces} traces, and \paperValue{agent-repairs/commits} of these repair commits correct the content (Appendix~\ref{app:editing}).
The cost is a median of \paperValue{agent-cost/seconds} seconds, \paperValue{agent-cost/turns} agent turns, and \paperValue{agent-cost/cost} USD per trace.

\subsection{Main Results}
\label{sec:main-results}

Table~\ref{tab:main} shows the competition-math results.
\mbta{} scores highest on average (\paperValue{main-ranking/a}), followed by Gandhi et al.\ (\paperValue{main-ranking/g}) and \mbti{} (\paperValue{main-ranking/i}), while \mbtr{}, whose training data are summarized, falls below Unedited SFT (\paperValue{rewrite-score/r} vs.\ \paperValue{rewrite-score/u}).
TI, which adds an instruction without training, lowers the Base model's score (\paperValue{ti-score/ti} vs.\ \paperValue{ti-score/base}).
The token-matched control scores about the same as Unedited SFT (\paperValue{token-control-score/control} vs.\ \paperValue{token-control-score/u}), so the \paperValue{extra-tokens/pct}\% additional tokens of \mbta{} alone do not explain its gain.
In three runs with different seeds, \mbta{} scores \paperValue{seed-ranges/amin} to \paperValue{seed-ranges/amax} and Unedited SFT \paperValue{seed-ranges/umin} to \paperValue{seed-ranges/umax} (Appendix~\ref{app:results}).
The three strategies rank in the same order as in the editing comparison of Section~\ref{sec:edit-compare} (\mbta{}, \mbti{}, then \mbtr{}).
The ASTRO condition separates position from content. It puts ASTRO's fixed phrases at the positions of \mbta{} except self-correction ones (Appendix~\ref{app:baselines}) and scores \paperValue{astro-score/score}, below Unedited SFT, which suggests that what the passages say matters, not only where they are.

\input{tables/general}

\paragraph{Untrained domains.}
\mbta{} scores above Unedited SFT on all five untrained benchmarks (Table~\ref{tab:general}).
The clearest gain is on LiveCodeBench (\paperValue{code-score/a} vs.\ \paperValue{code-score/u}), which suggests that behaviors inserted into math traces may carry over to code. The other gains are small (\paperValue{domain-gains/low} to \paperValue{domain-gains/high} points), and \mbti{} scores highest on GPQA and MuSiQue.

\input{figures/transfer}
\paragraph{Other students.}
\mbta{} also raises the average of MiniCPM5-2B-Base (\paperValue{minicpm-score/a} vs.\ \paperValue{minicpm-score/u} for Unedited SFT) and Granite 4.1 3B Base (\paperValue{granite-score/a} vs.\ \paperValue{granite-score/u}), though not on every benchmark (Appendix~\ref{app:results}).

%% file: figures/editing.tex
\begin{figure*}[t]
\centering
\includegraphics[width=\textwidth]{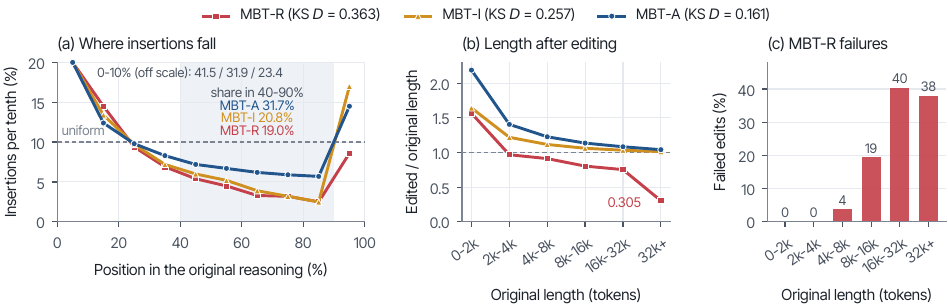}
\caption{(a) Share of insertions in each tenth of the original reasoning (axis capped at 20\%). \mbti{} places only \paperValue{editing-middle-share/mbt-i}\% of its insertions in the shaded middle, against \paperValue{editing-middle-share/mbt-a}\% for \mbta{} and 50\% under uniform placement. The legend gives the KS statistic $D$ against uniform placement (lower is more even). (b) Mean length ratio by original length. \mbtr{} shrinks long traces. (c) \mbtr{} fails more often on long traces.}
\label{fig:editing}
\end{figure*}

%% file: tables/main.tex
\begin{table}[t]
\centering
\footnotesize
\setlength{\tabcolsep}{2.4pt}
\begin{tabular}{@{}lcccccc@{}}
\toprule
 & \multicolumn{2}{c}{AIME} & \multicolumn{3}{c}{HMMT} & \\
\cmidrule(lr){2-3}\cmidrule(lr){4-6}
Method & 25 & 26 & Feb25 & Feb26 & Nov25 & Avg. \\
\midrule
Base model & 84.2 & 84.6 & 78.3 & 69.3 & 73.3 & 77.8 \\
Unedited SFT & \second{85.4} & 84.6 & 83.8 & 78.4 & 83.8 & 83.1 \\
\makecell[l]{Token-matched\\control} & 84.6 & 83.8 & \best{85.8} & 77.3 & 85.0 & 83.2 \\
\midrule
TI & 81.2 & 81.2 & 67.5 & 62.9 & 71.7 & 72.7 \\
ASTRO & 82.1 & 85.0 & 81.7 & 75.8 & 85.0 & 81.8 \\
CoT-Bridge & 82.9 & 83.3 & \second{85.0} & \second{79.5} & 85.4 & 83.2 \\
RefAug & 82.5 & 86.7 & \second{85.0} & 76.1 & 85.0 & 82.9 \\
Gandhi et al. & \best{87.5} & 85.8 & 84.2 & 78.0 & \second{85.8} & \second{84.2} \\
\midrule
\mbtr{} & 70.0 & 74.6 & 82.5 & 65.5 & 77.5 & 73.9 \\
\mbti{} & 82.9 & \second{87.5} & \second{85.0} & 78.8 & 85.4 & 83.8 \\
\rowcolor{cHl}\mbta{} & 85.0 & \best{90.0} & \second{85.0} & \best{79.9} & \best{87.5} & \best{85.4} \\
\bottomrule
\end{tabular}
\caption{Competition math with Qwen3.5-4B-Base (avg@8, \%). Avg.\ is over all \paperValue{table-math/problems} problems. TI is applied to the Base model without training. Bold and underline mark the best and second best per column.}
\label{tab:main}
\end{table}

%% file: tables/general.tex
\begin{table}[t]
\centering
\footnotesize
\setlength{\tabcolsep}{2.9pt}
\begin{tabular}{@{}lccccc@{}}
\toprule
Method & GPQA & LCB & MuSiQue & HotpotQA & 2Wiki \\
\midrule
Base model & 72.0 & 47.1 & 63.0 & 88.8 & 83.2 \\
Unedited SFT & 77.5 & 47.5 & 74.3 & \second{91.3} & 85.4 \\
\mbtr{} & 77.5 & 43.5 & 69.3 & 90.1 & 83.7 \\
\mbti{} & \best{79.2} & \second{48.0} & \best{74.8} & \second{91.3} & \second{85.7} \\
\rowcolor{cHl}\mbta{} & \second{78.1} & \best{50.3} & \second{74.6} & \best{91.4} & \best{86.0} \\
\midrule
Items & 198 & 131 & 2,417 & 7,405 & 12,576 \\
\bottomrule
\end{tabular}
\caption{Untrained domains (accuracy, \%), with avg@8 for GPQA Diamond and LiveCodeBench (LCB) and a single sample for multi-hop QA (2Wiki is 2WikiMultihopQA). Bold and underline mark the best and second best per column.}
\label{tab:general}
\end{table}

%% file: figures/transfer.tex
\begin{figure}[t]
\centering
\includegraphics[width=\columnwidth]{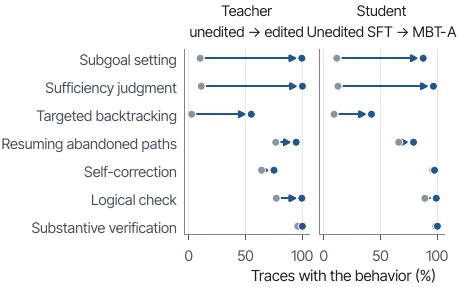}
\caption{Share of traces with each target behavior in \paperValue{transfer-caption/teacher} teacher traces before (gray) and after (blue) \mbta{} editing (left), and in \paperValue{student-transfer-caption/traces} traces each of the Unedited SFT (gray) and \mbta{} (blue) students (right).}
\label{fig:transfer}
\end{figure}

%% file: sections/discussion.tex
\section{Analysis}
\label{sec:analysis}

\subsection{Are the Behaviors Inserted?}
\label{sec:inserted}

We count the target behaviors with the LLM annotator under a \emph{target-behavior annotation scheme}, fixed before any trace is read. The annotator, chosen among five models by agreement with an agent's annotation (Appendix~\ref{app:teacher}), is told neither whether a trace is edited nor its condition. Two annotator runs agree with a per-behavior Cohen's kappa \citep{cohen1960} of \paperValue{target-reliability/tmin} to \paperValue{target-reliability/tmax} (teacher traces) and \paperValue{target-reliability/smin} to \paperValue{target-reliability/smax} (student traces).
On \paperValue{teacher-pair-counts/n} training traces (\paperValue{teacher-pair-counts/correct} correct, \paperValue{teacher-pair-counts/wrong} incorrect), all seven behaviors become more frequent after \mbta{} editing (Figure~\ref{fig:transfer}, left).
Originally rare behaviors rise the most, with subgoal setting from \paperValue{teacher-behaviour-rates/sb}\% to \paperValue{teacher-behaviour-rates/sa}\%, sufficiency judgment from \paperValue{teacher-behaviour-rates/cb}\% to \paperValue{teacher-behaviour-rates/ca}\%, and targeted backtracking from \paperValue{teacher-behaviour-rates/bb}\% to \paperValue{teacher-behaviour-rates/ba}\%, matching the commit history (Appendix~\ref{app:editing}).

\input{figures/forest}

\subsection{Do the Behaviors Transfer to the Student?}
\label{sec:transfer}

For each condition, we annotate \paperValue{student-counts/traces} student traces, two per problem on \paperValue{student-counts/problems} problems (one correct, one incorrect when possible).
Compared with the Unedited SFT student, the \mbta{} student shows the originally rare behaviors far more often, with subgoal setting in \paperValue{student-behaviour-rates/sa}\% vs.\ \paperValue{student-behaviour-rates/su}\% of traces, sufficiency judgment in \paperValue{student-behaviour-rates/ca}\% vs.\ \paperValue{student-behaviour-rates/cu}\%, and targeted backtracking in \paperValue{student-behaviour-rates/ba}\% vs.\ \paperValue{student-behaviour-rates/bu}\% (Figure~\ref{fig:transfer}, right).
For the four training-based prior methods and TI, the rates of the seven behaviors mostly stay near those of Unedited SFT (Appendix~\ref{app:behavior}).
In the example of Figure~\ref{fig:student-traces} (Appendix~\ref{app:behavior}), the Unedited SFT student takes one step too many to the 10th term and checks only its addition, whereas the \mbta{} student plans first, catches its own wrong 10th terms, and checks the total another way.

\subsection{Are the Behaviors Associated with Correctness?}
\label{sec:correctness}

Raw rates in correct and incorrect traces are confounded by difficulty, so we compare a correct and an incorrect trace of the same problem and condition (a matched pair) with conditional logistic regression.
Pooling the ten conditions gives \paperValue{matched-pairs/all} trace pairs, of which the \paperValue{matched-pairs/used} that differ in correctness enter the model
\begin{equation}
\operatorname{logit} \Pr(z_\tau = 1) = \alpha_{g(\tau)} + \beta\, h_\tau + \gamma \log L_\tau .
\end{equation}
Here $\tau$ is a student trace, $z_\tau = 1$ if it is correct, $h_\tau = 1$ if the behavior occurs in it at least once, $L_\tau$ is its length in characters, $\alpha_{g(\tau)}$ is an intercept for its pair $g(\tau)$, and $\operatorname{logit}(p) = \log \frac{p}{1-p}$.
Conditioning on each pair removes $\alpha_{g(\tau)}$, and with it problem difficulty and the training condition.

We fit one model per behavior with Firth's penalty \citep{firth1993}. The odds ratio $e^{\beta}$ is the factor by which the presence of the behavior multiplies the odds $\Pr(z_\tau = 1)/\Pr(z_\tau = 0)$ of a correct answer within a pair at a fixed length, and p-values are Holm-adjusted over the seven tests \citep{holm1979}.

Only the logical check is significantly associated with correctness (Figure~\ref{fig:forest}, odds ratio \paperValue{logical-check-association/odds}, adjusted $p = \paperValue{logical-check-association/p}$, Appendix~\ref{app:behavior}), in line with the finding that checks inside deductive steps distinguish correct traces \citep{chen2026anatomy}.
Substantive verification has a large but imprecise estimate (adjusted $p = \paperValue{verification-association/p}$) from only \paperValue{verification-association/n} discordant pairs.

\subsection{Does Deleting Metacognition Hurt?}
\label{sec:deletion}

To look beyond association, we edit in the opposite direction. If metacognitive sentences help, training without them should hurt more than training on equally few tokens of unedited traces.
The same editor model marks sentences of the original traces in eight metacognitive categories, which we delete while keeping solution steps and calculations (Appendix~\ref{app:deletion}). \emph{Delete-4} removes the four categories that correspond to our target behaviors and \emph{Delete-8} all eight.
A \emph{subsampled control} keeps fewer unedited traces to match the token count of Delete-8.

Delete-4 scores \paperValue{deletion-scores/d4} and Delete-8 \paperValue{deletion-scores/d8}, both below Unedited SFT (\paperValue{deletion-scores/u}) and even below the untrained Base model (\paperValue{deletion-scores/base}), whereas the subsampled control scores \paperValue{subsample-score/score} with only half the tokens (Appendix~\ref{app:deletion}).
Removing only the four categories that correspond to the target behaviors already costs \paperValue{deletion-drops/d4} points, more than half of the \paperValue{deletion-drops/d8}-point drop of Delete-8.
Deletion may also disturb coherence, but the metacognitive sentences, not the token count, appear to matter.

%% file: figures/forest.tex
\begin{figure}[t]
\centering
\includegraphics[width=\columnwidth]{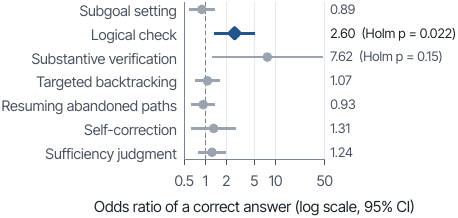}
\caption{Odds ratio of a correct answer when each behavior is present, within the same problem and condition (unadjusted 95\% CI). Only the logical check is significant after Holm correction.}
\label{fig:forest}
\end{figure}

%% file: sections/conclusion.tex
\section{Conclusion}
\label{sec:conclusion}

We propose MBT, which inserts seven metacognitive behaviors into teacher traces where the reasoning lacks them.
How the edits are made matters, not only what is inserted. Rewriting a whole trace at once summarizes long traces, and choosing all insertion points at once crowds them at the ends, while \mbta{}, which edits the trace as a version-controlled file line by line, avoids both.
The \mbta{} student outperforms Unedited SFT, a token-matched control, and five prior methods, \mbta{} also improves two other students, and the inserted behaviors clearly transfer. Deleting metacognitive sentences hurts far more than training on equally few unedited tokens.
A teacher trace's behaviors, not only its answers, are worth improving before distillation.

\section*{Limitations}

We use a single teacher model and train only on math problems. Whether the gains hold for other teachers and for interactive agentic tasks is untested, and the gains on untrained domains are mostly small.
Each condition except Unedited SFT and \mbta{} is run with a single seed, so differences of about one point between single runs should be read with caution.
\mbta{} is the most expensive strategy, with a median of \paperValue{conclusion-cost/turns} agent turns, \paperValue{conclusion-cost/seconds} seconds, and \paperValue{conclusion-cost/cost} USD per trace, and it needs an isolated execution environment.

Behaviors are counted by an LLM annotator chosen by its agreement with an agent's annotation under the undesirable-behavior scheme. We do not measure its agreement under the target-behavior scheme or with human annotators.
The behavior analyses show associations, not causes. The deletion experiment is closer to a causal test, but it removes several categories at once, so it does not isolate any single behavior.

\section*{Ethical Considerations}

This work uses publicly available math problems and benchmarks and involves no human participants.
The teacher, editor, judge, and annotator models are called through commercial APIs, and all requests and responses are logged.
We release the edited traces under the license of the source data (Appendix~\ref{app:release}).

\ifreviewversion\else
\section*{Acknowledgements}
This work was supported by Samsung Electronics Co., Ltd. (IO250624-13143-01), Institute of Information \& Communications Technology Planning \& Evaluation (IITP) grant funded by the Korea government (MSIT) [(No.2022-0-00959, RS-2022-II220959), NO.RS-2021-II211343, Artificial Intelligence Graduate School Program (Seoul National University)], the National Research Foundation of Korea (NRF) grant funded by the Korea government (MSIT) (No. 2022R1A3B1077720), and the BK21 FOUR program of the Education and Research Program for Future ICT Pioneers, Seoul National University in 2026. This work was also supported by the National AI Project of the Ministry of Science and ICT and the AIIS.
We use AI assistants for coding and for drafting and editing the manuscript, and the authors check all content.
\fi

%% file: sections/appendix.tex
\counterwithin*{table}{section}
\counterwithin*{figure}{section}
\renewcommand{\thetable}{\thesection\arabic{table}}
\renewcommand{\thefigure}{\thesection\arabic{figure}}

\section{Evidence from Prior Work}
\label{app:prior}

Table~\ref{tab:prior-evidence} collects the findings of prior work summarized in Section~\ref{sec:intro}, and the target behavior (Table~\ref{tab:target-behaviors}) that addresses each undesirable behavior.
The last row is not a behavior to insert but the reason for a rule that applies to every insertion, namely that a passage must state what it checked and what it found, so a bare reflection marker is not enough.

\begin{table*}[tp]
\centering
\footnotesize
\setlength{\tabcolsep}{4pt}
\renewcommand{\arraystretch}{1.1}
\begin{tabularx}{\textwidth}{@{}>{\raggedright\arraybackslash}p{2.9cm}>{\raggedright\arraybackslash}p{3.1cm}>{\raggedright\arraybackslash}X>{\raggedright\arraybackslash}p{3.0cm}@{}}
\toprule
Undesirable behavior & Evidence & Reported finding & Target behavior that addresses it \\
\midrule
No initial plan & \citet{bogdan2025thought}, \citet{marjanovic2025thoughtology} & When sentences are resampled one at a time, planning sentences are among those with the largest influence on the final answer. Models plan while executing instead of planning first. & subgoal setting \\
Missing logical check & \citet{chen2026anatomy} & Reflections that check the logic within a deduction are strongly associated with correct traces and much rarer in failed ones. Most reflections (78.8\%) occur in planning steps rather than deductive steps. & logical check \\
Superficial verification & \citet{chen2026anatomy}, \citet{stechly2024self} & Most reflections in failed traces are superficial verifications that cannot catch errors. When models verify and revise their own answers, performance drops, and revision helps only when a sound external verifier gives the verdicts. & substantive verification \\
Shallow backtracking & \citet{chen2026anatomy} & Most backtracking in failed traces jumps back less than a tenth of the trace, a local rewind rather than a return to an earlier premise. & targeted backtracking \\
Abandoning promising paths & \citet{wang2025underthinking}, \citet{lu2025retro} & Many incorrect answers contain a correct thought at least once. Training on traces revised to continue the abandoned paths makes traces shorter and accuracy higher. & resuming abandoned paths \\
Failure to self-correct & \citet{tsui2025selfcorrection}, \citet{tyen2024llms} & Errors that models correct when given as external input are largely left uncorrected in their own outputs. Fixing succeeds given the error location. & self-correction \\
Post-answer overthinking and unchecked final answer & \citet{marjanovic2025thoughtology} & Each problem has an optimal reasoning length. & sufficiency judgment \\
Reflection markers without an actual check & \citet{wang2025wait} & Preventing reflection markers such as ``Wait'' and ``Hmm'' during inference shortens traces while maintaining accuracy. & rule against insertions that consist only of markers \\
\bottomrule
\end{tabularx}
\caption{Undesirable behaviors in the traces of reasoning models reported by prior work, and the target behaviors that address them.}
\label{tab:prior-evidence}
\end{table*}

Table~\ref{tab:target-behaviors} lists the seven target behaviors, the undesirable behaviors each one addresses, and how often each already occurs in the original traces.

\input{tables/target_behaviors}

\section{Annotation of Teacher Traces}
\label{app:teacher}

\paragraph{Two annotation schemes.}
We use two annotation schemes.
The \emph{undesirable-behavior annotation scheme} defines \paperValue{diagnostic-codebook-size/types} undesirable behaviors (\paperValue{diagnostic-codebook-size/items} items including two sub-behaviors) and \paperValue{diagnostic-codebook-size/desirable} desirable behaviors, and is used only to find problems in teacher traces before editing (Figure~\ref{fig:teacher-behaviors} and Table~\ref{tab:undesirable-full}).
The \emph{target-behavior annotation scheme} defines the seven behaviors of Table~\ref{tab:target-behaviors} and is used to count them in teacher and student traces (Table~\ref{tab:target-behaviors}, Figure~\ref{fig:transfer}, and Appendix~\ref{app:behavior}).
The two schemes define behaviors differently and are applied to different trace sets, so similar-looking items can have different rates. For example, explicit planning in the first scheme (\paperValue{planning-schemes/correct}\% and \paperValue{planning-schemes/wrong}\% of correct and incorrect traces) and subgoal setting in the second (\paperValue{planning-schemes/subgoal}\%) are not the same item.
In both schemes, a behavior is counted as present if any part of the trace meets its definition.

\paragraph{Two sets of teacher traces.}
(1)~The \emph{diagnostic set} consists of \paperValue{diagnostic-counts/n} traces (\paperValue{diagnostic-counts/correct} correct, \paperValue{diagnostic-counts/wrong} incorrect) obtained by having the teacher solve \paperValue{diagnostic-counts/problems} of the \paperValue{diagnostic-counts/training} training problems four times each anew.
It serves to compare correct and incorrect traces on the same problems, and it is neither used for training nor edited (Figure~\ref{fig:teacher-behaviors}, Table~\ref{tab:undesirable-full}).
(2)~The \emph{before-after set} consists of \paperValue{before-after-counts/n} original training traces, \paperValue{before-after-counts/correct} correct and \paperValue{before-after-counts/wrong} incorrect according to the LLM judge, sampled in nearly equal numbers from each of the nine difficulty bins, with correct and incorrect traces length-matched within each bin, together with their \mbta{} edits (Table~\ref{tab:target-behaviors}, Figure~\ref{fig:transfer}).
The \paperValue{teacher-common-problems/n} problems of the diagnostic set are among the problems of the before-after set, and both sets underwent the same decontamination as the training problems.

\begin{figure*}[tp]
\centering
\includegraphics[width=\textwidth]{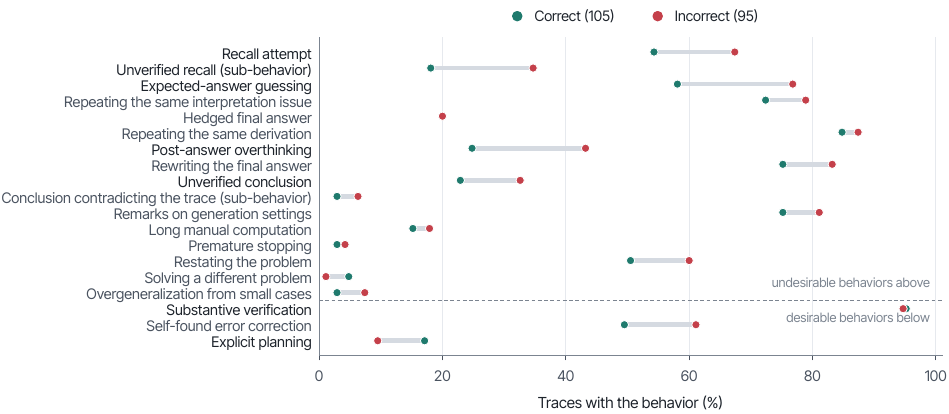}
\caption{All items of the undesirable-behavior annotation scheme on the \paperValue{diagnostic-caption/n} diagnostic teacher traces. Dark labels mark the items of Figure~\ref{fig:teacher-behaviors}.}
\label{fig:undesirable-full}
\end{figure*}

\paragraph{Full results.}
Table~\ref{tab:undesirable-full} gives every item of the undesirable-behavior annotation scheme on the diagnostic set, and Figure~\ref{fig:undesirable-full} plots the same rates.
Figure~\ref{fig:teacher-behaviors} shows the seven items related to the target behaviors, where ``no substantive verification'' and ``no initial plan'' are 100 minus the rates of the desirable behaviors substantive verification and explicit planning.
Because harder problems produce both more undesirable behaviors and more incorrect answers, the differences between correct and incorrect traces are partly confounded with difficulty.
The full scheme is in the released repository, and Appendix~\ref{app:behavior} describes the annotation prompt.

\input{tables/undesirable_full}

Figure~\ref{fig:teacher-trace} shows a teacher trace in which undesirable behaviors lead to a wrong answer.

\input{figures/teacher_trace}

\begin{figure}[tp]
\centering
\includegraphics[width=\linewidth]{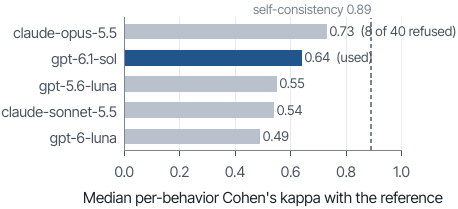}
\caption{Agreement of the candidate annotators with the reference annotation (Table~\ref{tab:annotator}). The dashed line is the agreement of gpt-6.1-sol with itself.}
\label{fig:annotator}
\end{figure}

\paragraph{Annotator selection.}
As a reference annotation, a claude-opus-5.5 agent reads each of \paperValue{reference-annotation-population/n} original teacher trace files in full and marks them with the undesirable-behavior annotation scheme, and we compare five candidate annotators with this reference on the same \paperValue{candidate-annotation-population/n} traces (Table~\ref{tab:annotator}).
Of the candidates, claude-opus-5.5 agrees best, but it refuses to annotate \paperValue{opus-annotation-failures/empty} of the \paperValue{opus-annotation-failures/total} reasoning traces and produces \paperValue{opus-annotation-failures/malformed} malformed outputs, so its agreement is computed on the remaining \paperValue{opus-valid-annotations/n} traces. Because the reference comes from the same model, this agreement may be inflated.
We therefore use gpt-6.1-sol, which agrees second best without refusals.
Figure~\ref{fig:annotator} shows the agreement of each candidate. To see how much its annotations vary between calls, we annotate the \paperValue{repeated-annotation-count/n} traces a second time with gpt-6.1-sol and compare the two annotations.

\begin{table}[tp]
\input{tables/panel_annotator}
\end{table}

\section{Editing Instructions and the \mbta{} Environment}
\label{app:instructions}

\begin{figure*}[tp]
\centering
\includegraphics[width=\textwidth]{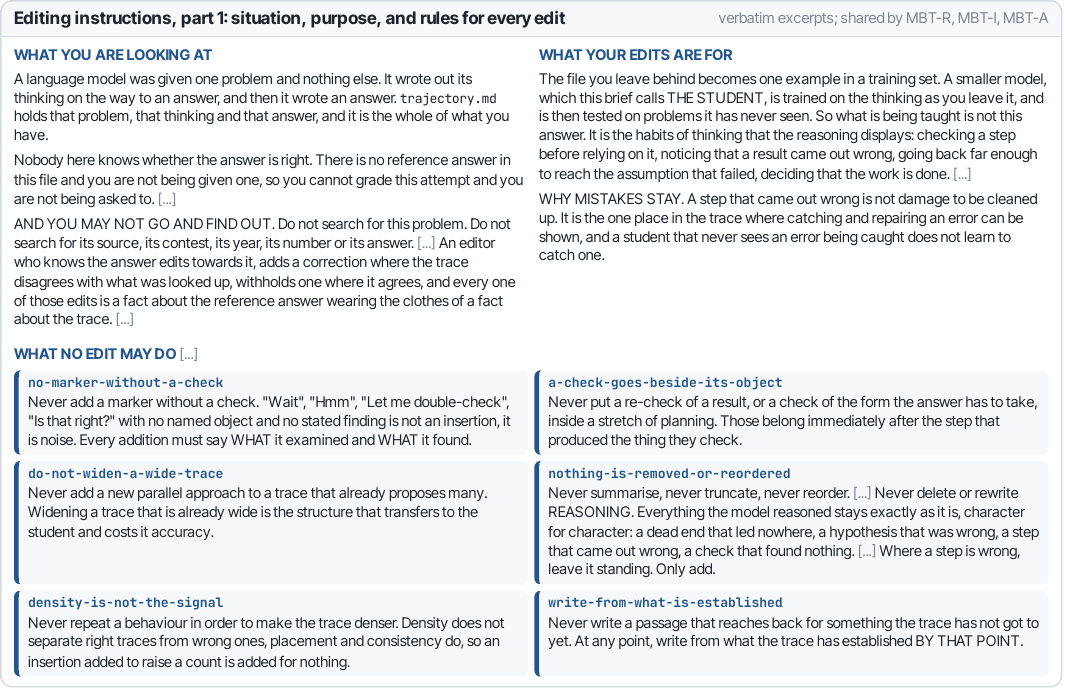}
\caption{Editing instructions, part 1 (verbatim excerpts), covering the situation, the purpose of the edits, and the rules that apply to every edit.}
\label{fig:prompt-editing-1}
\end{figure*}

\begin{figure*}[tp]
\centering
\includegraphics[width=\textwidth]{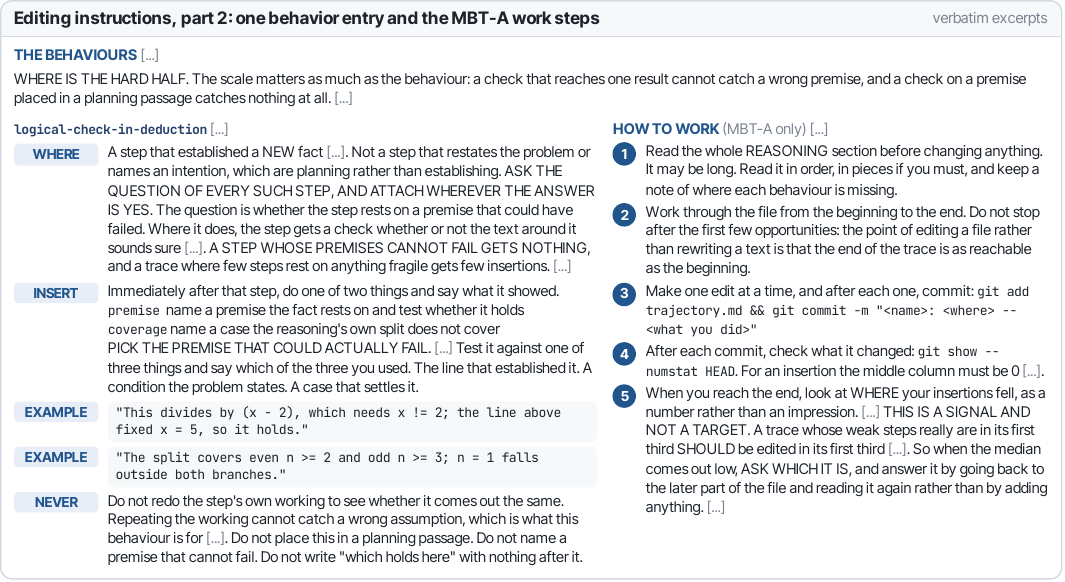}
\caption{Editing instructions, part 2 (verbatim excerpts), showing the logical-check entry and the work steps.}
\label{fig:prompt-editing-2}
\end{figure*}

\paragraph{Shared instructions.}
The three strategies use the same instructions, which first describe the situation (the trace comes without a reference answer, and the problem and its answer must not be looked up) and the purpose (the edited trace becomes training data, so it teaches habits of thinking, not the answer).
They then give named rules for every edit (e.g., \texttt{no-marker-without-a-check}) and one entry per behavior, consisting of where to insert (WHERE), what to insert (INSERT), examples (EXAMPLE), and prohibitions (NEVER).
For \mbta{}, numbered work steps follow, including the review of insertion positions.
A behavior with no position that meets its condition receives nothing, and \mbta{} records this as an empty commit with the reason.
Figures~\ref{fig:prompt-editing-1} and~\ref{fig:prompt-editing-2} reproduce central parts verbatim, with omissions marked by [...]. The full text is in the released repository.

\begin{figure*}[tp]
\centering
\includegraphics[width=\textwidth]{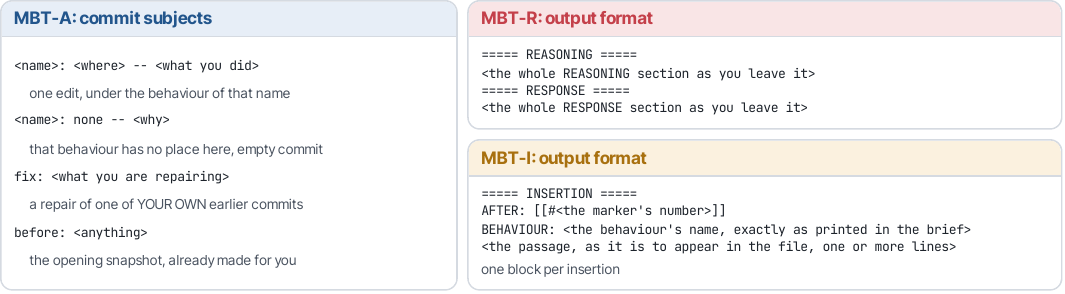}
\caption{Strategy-specific formats, namely the four commit subjects of \mbta{}, the output of \mbtr{}, which rewrites the whole trace, and the output of \mbti{}, one block per insertion.}
\label{fig:prompt-formats}
\end{figure*}

\paragraph{Strategy-specific parts.}
For \mbta{}, the full instructions are the first user message to the agent, and the working directory contains only the version history (\texttt{.git}) and \texttt{trajectory.md}.
For \mbtr{}, the instructions without the parts on the repository, editing tools, and commits are preceded by a note that the task is done in a single answer and followed by the entire trace file. The editor's output rewrites the reasoning and response sections in full.
\mbti{} uses the same instructions as \mbtr{}, shows the reasoning section with a \texttt{[[\#k]]} marker after each paragraph, and asks for a list of insertions instead of the whole trace.
Figure~\ref{fig:prompt-formats} shows the output format of each strategy.

\paragraph{\mbta{} execution environment.}
The harness is Claude Code 2.1.283, which calls the editor model (DeepSeek-V4.1-Flash, reasoning effort max) with an output limit of 128,000 tokens per request, up to 10 retries per request, and automatic context compaction at 917,504 tokens. The harness sends no temperature or top\_p.
The agent runs in an isolated E2B sandbox \citep{e2b2026} with 1 CPU and 2 GiB of memory.
One agent run may take at most 3,600 seconds, and the sandbox lives 7,200 seconds. No trace reaches the time limit.

\section{Experimental Setup Details}
\label{app:setup}

\paragraph{Training data.}
All problems, traces, and evaluation benchmarks are in English.
Each condition trains on the traces that are edited successfully and fit within the maximum student length of \paperValue{max-student-length/length} tokens (Table~\ref{tab:train-data}).
For decontamination, we use seven math benchmarks besides the evaluation benchmarks, namely MATH-500 \citep{lightman2024verify}, AMC 2023, AIME 2024, HMMT February 2023, HMMT February 2024, OlympiadBench \citep{he2024olympiadbench}, and GSM8K \citep{cobbe2021gsm8k}.
The \paperValue{control-extra/extra} problems of the token-matched control are sampled from the same deduplicated and decontaminated pool in the same way as the \paperValue{control-training-population/n} problems (by difficulty bin), do not overlap with them, and each adds one unedited trace.

\paragraph{Training.}
All students and conditions use the hyperparameters of Table~\ref{tab:hyper}.
We search the learning rate over $\{3 \times 10^{-6}, 10^{-5}, 3 \times 10^{-5}\}$ and the number of epochs over $\{5, 10, 15\}$ and choose the setting with the best score of Unedited SFT on the \paperValue{search-problem-count/n} competition-math problems of the evaluation (Section~\ref{sec:setup}). This setting, a learning rate of $\paperValue{selected-training-recipe/lr}$ and \paperValue{selected-training-recipe/epochs} epochs, is applied unchanged to every other condition. The other hyperparameters follow the technical reports of recent open models.
Training and evaluation run on one node with eight NVIDIA H100 GPUs for about four weeks, about 5,400 GPU hours in total.
The judge and annotator models are called through APIs, and their sizes are not public. The teacher and editor, DeepSeek-V4.1-Flash, has open weights, but we also call it through its API.
All comparisons use the checkpoint after \paperValue{setup-checkpoint-repeated/epoch} epochs. Since the learning rate is constant after warmup, this checkpoint does not depend on the length of the schedule. Table~\ref{tab:software} lists the software and model call settings.

\paragraph{Evaluation.}
For math, we draw eight samples for each problem. For the other domains, we draw one sample per item on MuSiQue, HotpotQA, and 2WikiMultihopQA and eight on LiveCodeBench v6 and GPQA. LiveCodeBench v6 consists of the \paperValue{livecodebench-release-size/n} problems of release\_v6 whose contest dates fall between February 1 and May 1, 2025.
Generation uses max\_new\_tokens \paperValue{generation-hyperparameters/output}, a context window of \paperValue{generation-hyperparameters/context}, temperature \paperValue{generation-hyperparameters/temp}, top\_p \paperValue{generation-hyperparameters/p}, top\_k $\paperValue{generation-hyperparameters/k}$, and thinking enabled, served with vLLM \citep{kwon2023vllm}.
An output from which no answer can be extracted (one that exhausts the generation limit, stops without writing anything, or lacks the answer format) is regenerated, up to \paperValue{appendix-answer-retries/n} times per sample. A sample that hits the output limit five times in a row or yields no extractable answer within \paperValue{generation-retries/tries} tries is scored 0.

\begin{figure}[tp]
\centering
\includegraphics[width=\columnwidth]{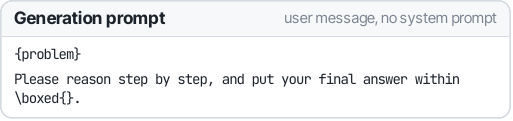}
\caption{Prompt for teacher generation and for the math evaluation of the students.}
\label{fig:prompt-generation}
\end{figure}

Math, multi-hop QA, and GPQA are graded by the LLM judge, and LiveCodeBench by running the tests.
The generation prompt (Figure~\ref{fig:prompt-generation}) is used both for teacher generation and for the math evaluation of the students, and the grading prompts come from public evaluation code (Table~\ref{tab:judge-prompts}).

\paragraph{Judge reliability.}
On all \paperValue{judge-problem-count/n} competition-math problems, we compare the verdicts of the LLM judge (gpt-5.6-luna) with those of math\_verify on \paperValue{judge-agreement-cohort/samples} samples per training condition (four evaluations of eight samples per problem) and \paperValue{judge-agreement-cohort/ti-samples} for TI (two evaluations). They agree on \paperValue{judge-agreement-range/low}\% to \paperValue{judge-agreement-range/high}\% of samples per condition.

\begin{table}[tp]
\input{tables/panel_train-data}
\end{table}

\begin{table}[tp]
\centering
\footnotesize
\setlength{\tabcolsep}{4pt}
\begin{tabularx}{\linewidth}{@{}>{\raggedright\arraybackslash}p{1.9cm}>{\raggedright\arraybackslash}X@{}}
\toprule
Use & Source \\
\midrule
Competition math & NVIDIA NeMo-Skills \citep{nvidia2026nemoskills}, \texttt{judge/math.yaml} \\
GPQA Diamond & Evaluation code of Humanity's Last Exam \citep{phan2025hle}, centerforaisafety/hle commit 22ed3074, \texttt{JUDGE\_PROMPT} in \texttt{hle\_eval/run\_judge\_results.py} \\
MuSiQue, HotpotQA, 2WikiMultihopQA & Grader template of SimpleQA Verified \citep{haas2025simpleqa}, \nolinkurl{SIMPLEQA_VERIFIED_GRADER_TEMPLATE} in \nolinkurl{src/inspect_evals/simpleqa/grader_templates.py}, UKGovernmentBEIS/inspect\_evals commit dbd3dd25 \\
\bottomrule
\end{tabularx}
\caption{Sources of the grading prompts, each sent as the user message to the LLM judge.}
\label{tab:judge-prompts}
\end{table}

\begin{table}[tp]
\centering
\footnotesize
\setlength{\tabcolsep}{4pt}
\begin{tabularx}{\linewidth}{@{}l>{\raggedright\arraybackslash}X@{}}
\toprule
Item & Value \\
\midrule
Framework & TRL \citep{vonwerra2020trl}, Megatron, accelerate \\
Optimizer & Muon \citep{jordan2024muon,liu2025muon} with AdamW, betas \paperValue{optimizer-hyperparameters/beta1} and \paperValue{optimizer-hyperparameters/beta2}, eps \paperValue{optimizer-hyperparameters/eps} \\
Learning rate & \paperValue{learning-rate-warmup/rate}, constant with \paperValue{learning-rate-warmup/warmup} warmup steps \\
Batch size & \paperValue{batch-size/batch} \\
Epochs & \paperValue{epochs-table/epochs} \\
Precision & bf16, fp32 master weights \\
Weight decay, grad clip & \paperValue{clip-decay/decay}, \paperValue{clip-decay/clip} \\
Maximum length & \paperValue{max-length-table/length} (longer examples excluded), no packing, padding-free \\
Loss & assistant tokens only, averaged per token (Eq.~\ref{eq:loss}) \\
\bottomrule
\end{tabularx}
\caption{Training hyperparameters, shared by all students and conditions.}
\label{tab:hyper}
\end{table}

\begin{table}[tp]
\centering
\footnotesize
\setlength{\tabcolsep}{4pt}
\begin{tabularx}{\linewidth}{@{}l>{\raggedright\arraybackslash}X@{}}
\toprule
Component & Version or setting \\
\midrule
Training & Python \paperValue{training-software/python}, torch \paperValue{training-software/torch}, transformers \paperValue{training-software/transformers}, trl \paperValue{training-software/trl}, accelerate \paperValue{training-software/accelerate}, liger-kernel \paperValue{training-software/liger}, triton \paperValue{training-software/triton} \\
Serving & vllm \paperValue{serving-software/vllm}, torch \paperValue{serving-software/torch}, flashinfer-python \paperValue{serving-software/flashinfer} \\
Editing harness & Claude Code 2.1.283, E2B sandbox \\
LLM judge & gpt-5.6-luna, reasoning effort medium \\
Annotator & gpt-6.1-sol, reasoning effort medium \\
Teacher & DeepSeek-V4.1-Flash, reasoning effort max, temperature 1.0, top\_p 0.95, one sample per training problem \\
Editor & DeepSeek-V4.1-Flash, reasoning effort max. \mbtr{} and \mbti{} use temperature 1.0, top\_p 0.95, and max\_tokens 131,072, and \mbta{} uses the harness settings \\
\bottomrule
\end{tabularx}
\caption{Software versions and model call settings.}
\label{tab:software}
\end{table}

\paragraph{Released materials.}\label{app:release}
For the \paperValue{release-problem-count/n} Nemotron-Math-v2 problems, we release the original teacher traces and the traces edited by \mbtr{}, \mbti{}, and \mbta{}, together with the code, the full text of every prompt summarized in this appendix, and both annotation schemes, at \url{\repourl}.
Following the license of Nemotron-Math-v2, the dataset is released under CC BY-SA 4.0.
The released data contain only Nemotron-Math-v2 problems and outputs of DeepSeek-V4.1-Flash, whose terms allow using outputs to train other models.
The outputs of the OpenAI judge and annotator are used only for evaluation and analysis, not as training data, and we do not redistribute the evaluation benchmarks.
Table~\ref{tab:licenses} lists the licenses or terms of the main artifacts we use.

\begin{table*}[tp]
\centering
\footnotesize
\setlength{\tabcolsep}{4pt}
\renewcommand{\arraystretch}{1.06}
\begin{tabularx}{\textwidth}{@{}>{\raggedright\arraybackslash}p{4.4cm}>{\raggedright\arraybackslash}p{3.6cm}>{\raggedright\arraybackslash}X@{}}
\toprule
Artifact & Use & License or terms \\
\midrule
\tgroup{3}{Data}
Nemotron-Math-v2 & training problems & CC BY 4.0 (AoPS subset), CC BY-SA 4.0 (StackExchange subsets) \\
MathArena AIME 2025, AIME 2026, HMMT Feb 2025, Feb 2026, Nov 2025 & evaluation & CC BY-NC-SA 4.0 \\
GPQA Diamond & evaluation & CC BY 4.0, examples not to be shown online \\
LiveCodeBench v6 & evaluation & CC (data), MIT (code) \\
MuSiQue & evaluation & CC BY 4.0 \\
HotpotQA & evaluation & CC BY-SA 4.0 \\
2WikiMultihopQA & evaluation & Apache-2.0 \\
\midrule
\tgroup{3}{Models}
DeepSeek-V4.1-Flash & teacher, editor & MIT (weights), and the API terms allow training other models on outputs \\
Qwen3.5-4B-Base, MiniCPM5-2B-Base, Granite 4.1 3B Base & students & Apache-2.0 \\
gpt-oss-120b & difficulty bins & Apache-2.0 \\
gpt-5.6-luna, gpt-6.1-sol (API) & judge, annotator & OpenAI Services Agreement \\
\midrule
\tgroup{3}{Software}
TRL, vLLM, Math-Verify, NeMo-Skills, E2B SDK & training, serving, grading, sandbox & Apache-2.0 \\
HLE evaluation code, inspect\_evals, Muon & grading prompts, optimizer & MIT \\
Claude Code & agent harness & Anthropic Commercial Terms of Service \\
\bottomrule
\end{tabularx}
\caption{Licenses and terms of the artifacts used in this work, checked on the official pages on October 12, 2026.}
\label{tab:licenses}
\end{table*}

\section{Reimplementation of Prior Methods}
\label{app:baselines}

Table~\ref{tab:baselines} summarizes how we reimplement each prior method.
Every model call uses DeepSeek-V4.1-Flash with the original paper's prompt, adapted to math only where Table~\ref{tab:baselines} says so.

\paragraph{ASTRO.}
The original paper runs Monte Carlo tree search with Llama-3.3-70B and, while linearizing the search tree into one solution, inserts fixed phrases.
Where the search returns to an earlier node, it inserts ``But wait, are we solving the problem correctly so far? Hmm...'' and ``I'm not sure if we're solving the problem correctly so far.'', after which the model writes, with a few-shot prompt, ``Let's go back to where we ...'' (an earlier node) or ``Let's restart our solution from the beginning.'' (the root).
Between nodes that proceed well and after the last node, it inserts the same reflection phrase and ``Our solution seems to be correct so far.''

\begin{figure*}[tp]
\centering
\includegraphics[width=\textwidth]{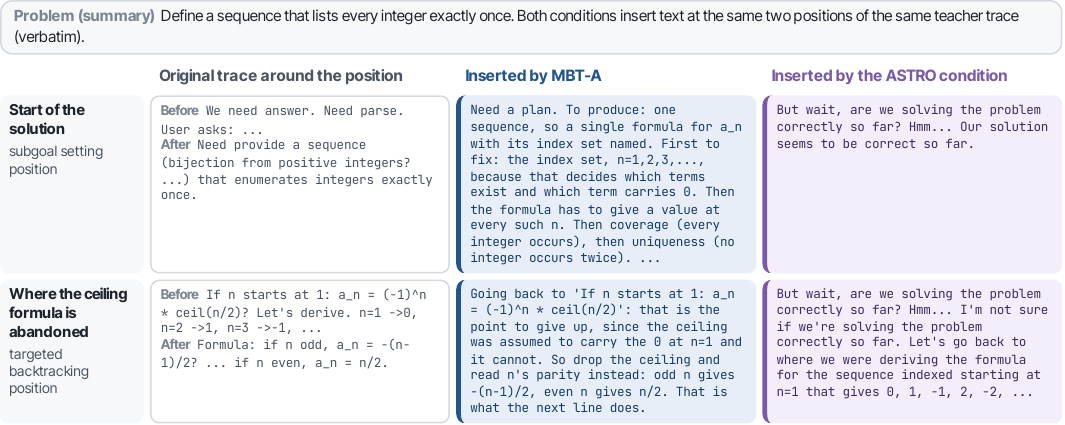}
\caption{Text inserted by \mbta{} and by the ASTRO condition at the same two positions of the same teacher trace (verbatim). ASTRO's fixed phrases do not state what was checked.}
\label{fig:astro-example}
\end{figure*}

Our traces have no search tree, so we replay the \mbta{} commits one by one to read the position and behavior of each inserted passage, and insert the original phrases at those positions instead of the \mbta{} passages. The blank after ``Let's go back to where we'' is filled by the editor model with the original few-shot prompt.
The ASTRO condition thus intervenes at the same positions as \mbta{}, but each intervention consists only of the original phrases and the short clause that fills their blank (Figure~\ref{fig:astro-example}).
It inserts nothing at self-correction positions, because the original paper has no phrase for continuing an incorrect node. In the example, \mbta{} also inserts a self-correction passage right before the second position (a recomputation showing that the ceiling formula gives $-1$ at $n = 1$).

\input{tables/baselines}

\section{Editing Statistics}
\label{app:editing}

\begin{figure}[tp]
\centering
\includegraphics[width=\linewidth]{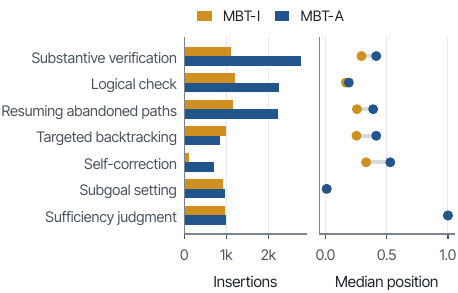}
\caption{Insertions per behavior, with the count (left) and the median position in the original reasoning (right).}
\label{fig:per-behavior}
\end{figure}

\paragraph{Measurement.}
Length ratios are computed over the successful edits, which for \mbtr{} are its \paperValue{edit-measurements/edits} edits and their originals.
\mbtr{} rewrites the whole trace, so we count its insertions as newly added chunks of at least 150 characters in a line-level comparison with the original.
\mbti{} positions follow from the markers, and \mbta{} positions are mapped back to the original trace by replaying the changes of each commit.
For the $M$ insertions of a strategy pooled over traces, with positions $\pi_1, \dots, \pi_M$, the Kolmogorov-Smirnov statistic against the uniform distribution is
\begin{equation}
D = \sup_{u \in [0,1]} \bigl| \hat{F}(u) - u \bigr|,
\end{equation}
where $\hat{F}(u) = \frac{1}{M} \sum_{l=1}^{M} \mathbf{1}[\pi_l \le u]$ is the empirical distribution function of the positions and $\mathbf{1}[\cdot]$ is 1 if its condition holds and 0 otherwise. Smaller values mean more even placement.
Table~\ref{tab:editing-full} gives all statistics, and Figure~\ref{fig:per-behavior} the counts and median positions per behavior.

\begin{figure}[tp]
\centering
\includegraphics[width=\linewidth]{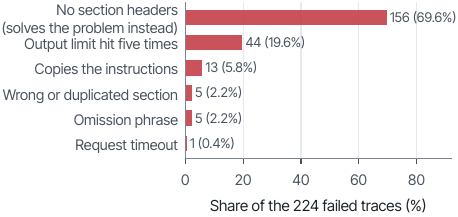}
\caption{Why \mbtr{} fails. Most failed outputs solve the problem instead of returning the edited trace.}
\label{fig:mbtr-failures}
\end{figure}

\paragraph{\mbtr{} failures.}
Table~\ref{tab:mbtr-failures} and Figure~\ref{fig:mbtr-failures} classify the \paperValue{failure-count-appendix/n} failures by the reason of the last call.
Outputs without section headers are mostly cases in which the editor copies the trace into its own thinking and then continues solving the problem.
The failure rate is \paperValue{failure-buckets/a}\% below 4k tokens, \paperValue{failure-buckets/b}\% for 4k to 8k, \paperValue{failure-buckets/c}\% for 8k to 16k, \paperValue{failure-buckets/d}\% for 16k to 32k, and \paperValue{failure-buckets/e}\% above 32k.

\paragraph{\mbta{} workload and commits.}
Table~\ref{tab:mbta-cost} gives the work per trace, with cost at the official DeepSeek API prices at the time of the runs.
The \paperValue{commits-appendix/total} commits consist of \paperValue{commits-appendix/initial} initial snapshots, \paperValue{commits-appendix/insertions} behavior insertions, \paperValue{commits-appendix/repairs} repairs of the agent's own earlier insertions, and \paperValue{commits-appendix/empty} empty commits for behaviors with no valid position.
Most repairs polish the sentences of an inserted passage, splitting long sentences into short ones like those of the original trace, but \paperValue{content-repairs-appendix/n} correct the content of the passage (Table~\ref{tab:fix-commits}). In one case, the agent recomputes with Python the rounding in a self-check that it inserted six commits earlier and corrects it.
Per behavior (Figure~\ref{fig:per-behavior}), \mbta{} inserts more of every behavior than \mbti{} except targeted backtracking (\paperValue{backtracking-insertions/a} vs.\ \paperValue{backtracking-insertions/i}), most of all substantive verification, logical check, resuming abandoned paths, and self-correction, and places five of the seven behaviors later in the trace (the other two tie).

\begin{table}[tp]
\input{tables/panel_editing-full}
\end{table}

\begin{table}[tp]
\input{tables/panel_mbtr-failures}
\end{table}

\begin{table}[tp]
\input{tables/panel_mbta-cost}
\end{table}

\begin{table}[tp]
\input{tables/panel_fix-commits}
\end{table}

\section{Additional Results}
\label{app:results}

\begin{figure*}[tp]
\centering
\includegraphics[width=\textwidth]{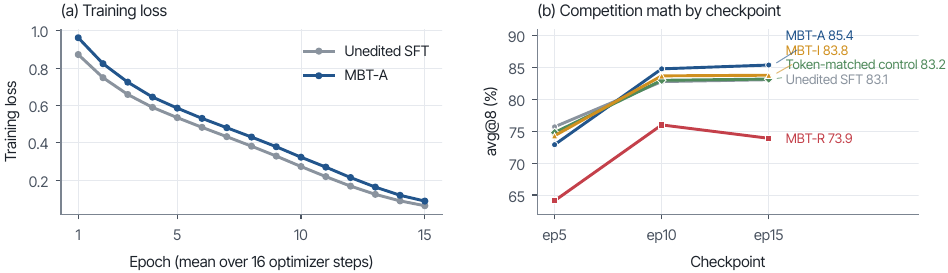}
\caption{(a) Training loss of the main Unedited SFT and \mbta{} runs, averaged per epoch. (b) Math score (avg@\paperValue{curve-rollouts/n} on the \paperValue{figure-math-size/n} problems) at the three evaluated checkpoints.}
\label{fig:curves}
\end{figure*}

\begin{figure}[tp]
\centering
\includegraphics[width=\linewidth]{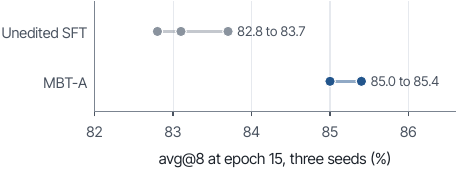}
\caption{Final math scores of the three seeds of Unedited SFT and \mbta{}.}
\label{fig:seeds}
\end{figure}

\paragraph{Checkpoints and seeds.}
Table~\ref{tab:epochs} gives the math score of every condition after \paperValue{epoch-table-checkpoints/first}, \paperValue{epoch-table-checkpoints/second}, and \paperValue{epoch-table-checkpoints/last} epochs, and Figure~\ref{fig:curves} the training loss of the main runs and the scores of five conditions.
The \mbta{} run has a higher training loss throughout and a lower score at epoch \paperValue{epoch-crossover/early}, but overtakes Unedited SFT by epoch \paperValue{epoch-crossover/later}.
Unedited SFT and \mbta{} are each trained three times with different seeds (Table~\ref{tab:seeds}). At epoch \paperValue{seed-comparison-epoch/n}, every \mbta{} run scores higher than every Unedited SFT run (\paperValue{seed-ranges-appendix/amin} to \paperValue{seed-ranges-appendix/amax} vs.\ \paperValue{seed-ranges-appendix/umin} to \paperValue{seed-ranges-appendix/umax}, Figure~\ref{fig:seeds}).

\begin{figure*}[tp]
\centering
\includegraphics[width=\textwidth]{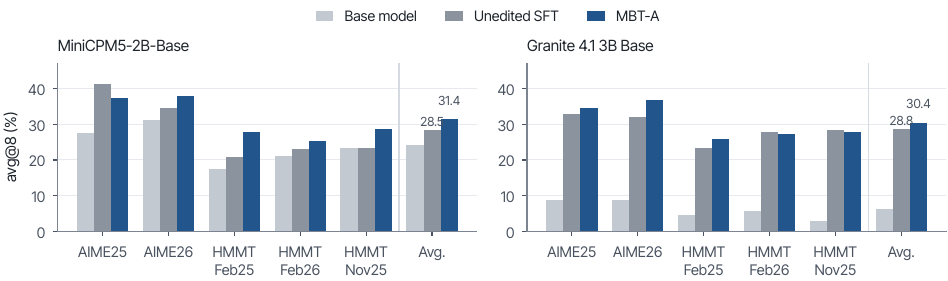}
\caption{Competition math for the other students (Table~\ref{tab:other-students}). \mbta{} raises the average of both models, though not every benchmark.}
\label{fig:other-students}
\end{figure*}

\paragraph{Other students.}
Table~\ref{tab:other-students} and Figure~\ref{fig:other-students} give the per-benchmark results of MiniCPM5-2B-Base and Granite 4.1 3B Base.
\mbta{} improves the average of both models, but not every benchmark, since Unedited SFT is higher on AIME 2025 for MiniCPM5 and on HMMT February 2026 and November 2025 for Granite.

\begin{table}[tp]
\input{tables/panel_epochs}
\end{table}

\begin{table}[tp]
\input{tables/panel_other-students}
\end{table}

\begin{table}[tp]
\input{tables/panel_seeds}
\end{table}

\section{Behavior Analysis Details}
\label{app:behavior}

\begin{figure*}[tp]
\centering
\includegraphics[width=\textwidth]{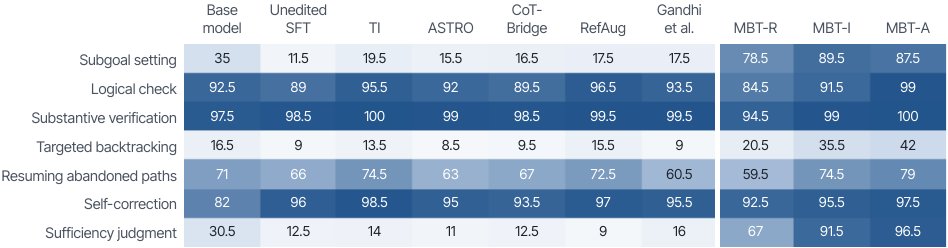}
\caption{Share (\%) of \paperValue{student-behaviour-caption/n} student traces in which each target behavior occurs, for all ten conditions. The three MBT strategies (right of the white line) raise the behaviors that are rare under the other conditions.}
\label{fig:all-behaviors}
\end{figure*}

\begin{figure*}[tp]
\centering
\includegraphics[width=\textwidth]{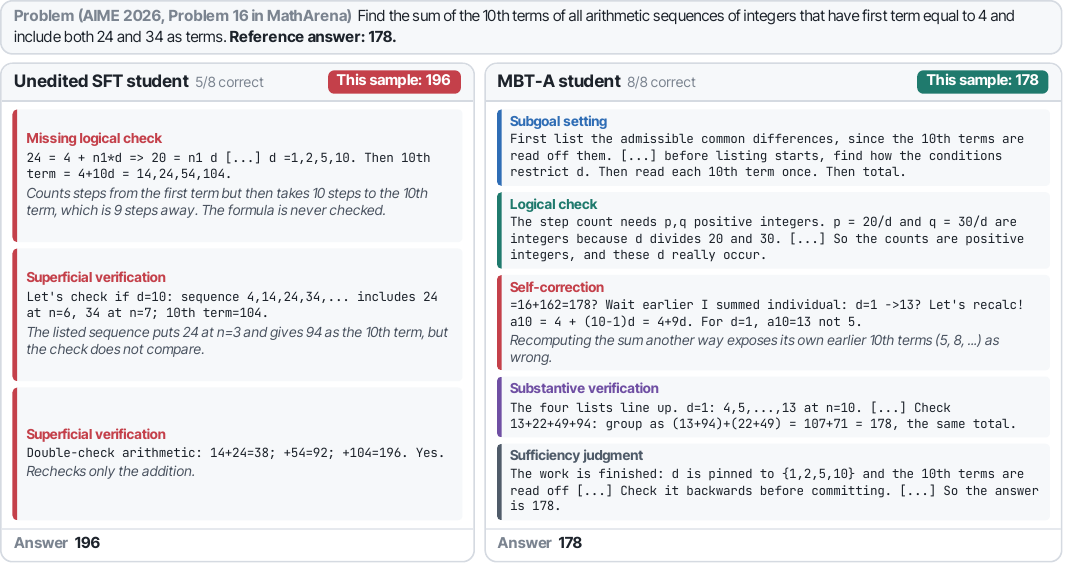}
\caption{Two students on AIME 2026, Problem 16 in MathArena (verbatim excerpts, labels and notes by the authors). The Unedited SFT student uses a wrong formula for the 10th term and checks only its addition. The \mbta{} student plans, checks that every candidate difference works, fixes its earlier 10th terms, and rechecks its sum by regrouping.}
\label{fig:student-traces}
\end{figure*}

\input{tables/clogit}

\paragraph{Annotation protocol.}
We annotate ten conditions, namely the Base model, Unedited SFT, \mbtr{}, \mbti{}, \mbta{}, the four training-based prior methods, and TI.
We draw \paperValue{annotation-problem-sample/selected} of the \paperValue{annotation-problem-sample/total} competition-math problems at random (fixed seed) and take two traces per problem for every condition, one correct and one incorrect when both exist, otherwise two of one kind.

For each trace, the annotator (gpt-6.1-sol, reasoning effort medium) receives as the user message the annotation instructions, the annotation scheme, and the trace (problem, reasoning, and final answer) with numbered lines.
It answers in a fixed JSON format with, for each behavior, its presence, the number of occurrences, and the starting lines. It does not see the correctness of the answer or the training condition.
The target-behavior annotation scheme spells out the definitions of Table~\ref{tab:target-behaviors}.

\paragraph{All conditions.}
Figure~\ref{fig:all-behaviors} gives the rates of the seven behaviors for all ten conditions.
The four training-based prior methods and TI stay close to Unedited SFT on the rare behaviors (subgoal setting, targeted backtracking, and sufficiency judgment), whereas all three MBT strategies raise them. \mbta{} has the highest rates of logical check and targeted backtracking.
Figure~\ref{fig:student-traces} compares the two students of Section~\ref{sec:transfer} on one problem.
Unedited SFT lowers some rates relative to the Base model, for example subgoal setting from \paperValue{base-student-rates/sb}\% to \paperValue{base-student-rates/su}\% and sufficiency judgment from \paperValue{base-student-rates/cb}\% to \paperValue{base-student-rates/cu}\%.

\paragraph{Association with correctness.}
We fit the conditional logistic regression with Firth's penalty, using our own NumPy implementation, which we release with the code.
Table~\ref{tab:clogit} gives the estimates of the conditional logistic regression of Section~\ref{sec:correctness}, which Figure~\ref{fig:forest} plots, with Wald 95\% confidence intervals.
Among the \paperValue{clogit-pairs-appendix/n} pairs, a pair is discordant in a behavior if the behavior is present in one trace and absent in the other. Only such pairs carry direct information about $\beta$, so a behavior with few discordant pairs, such as substantive verification (\paperValue{verification-discordant/n}), has a wide confidence interval.

\section{Deletion Conditions}
\label{app:deletion}

\begin{table}[tp]
\centering
\footnotesize
\setlength{\tabcolsep}{4pt}
\renewcommand{\arraystretch}{1.08}
\begin{tabularx}{\columnwidth}{@{}>{\raggedright\arraybackslash}p{\ifreviewversion 1.8cm\else 2.05cm\fi}>{\raggedright\arraybackslash}X@{}}
\toprule
Category & Text deleted (target behavior) \\
\midrule
Problem understanding & Restating the problem, listing conditions, settling the meaning of the question, and remarks about the request itself such as the answer format or length \\
Planning and decom\-position$^{*}$ & Saying what to do before doing it, subgoals, previews of the next step, and plans for writing the final answer (\textit{subgoal setting}) \\
Strategy selection and justification & Listing methods, choosing one, and justifying the choice. The solution by the chosen method is kept \\
Progress monitoring & Summaries of what has been obtained and what remains. Judgments that the problem is hard or one is stuck \\
Error detection and correction$^{*}$ & (a) An incorrect step together with the text that catches it, leaving only the corrected solution. (b) Sentences saying that a path is abandoned or that one goes back, keeping the abandoned path's work (\textit{self-correction, targeted backtracking}) \\
Verification$^{*}$ & All text that rechecks a result already obtained, such as substitution, recomputation, another method, special cases, premise checks, and verdicts such as ``Yes.'' (\textit{substantive verification, logical check}) \\
Sufficiency judgment$^{*}$ & Judgments that the work is enough to answer or that more is needed (\textit{sufficiency judgment}) \\
Reflection and evaluation & Looking back on and evaluating the solution as a whole \\
\bottomrule
\end{tabularx}
\caption{The eight deletion categories, with the corresponding target behaviors in parentheses. Delete-4 removes the four categories marked with $^{*}$, and Delete-8 removes all eight.}
\label{tab:delete-categories}
\end{table}

\ifreviewversion
\begin{table}[tp]
\input{tables/panel_deletion}
\end{table}

\begin{figure}[tp]
\centering
\includegraphics[width=\linewidth]{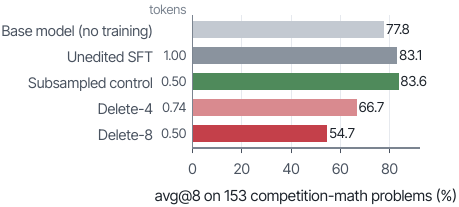}
\caption{Training on traces with metacognitive sentences deleted. The tokens column gives the training tokens relative to the original data. The subsampled control has as many tokens as Delete-8.}
\label{fig:deletion}
\end{figure}
\else
\begin{figure*}[tp]
\begin{minipage}[b]{0.47\textwidth}
\setcaptype{table}
\input{tables/panel_deletion}
\end{minipage}\hfill
\begin{minipage}[b]{0.5\textwidth}
\centering
\includegraphics{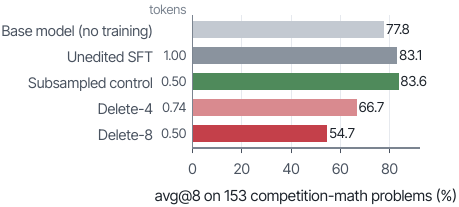}
\caption{Training on traces with metacognitive sentences deleted. The tokens column gives the training tokens relative to the original data. The subsampled control has as many tokens as Delete-8.}
\label{fig:deletion}
\end{minipage}
\end{figure*}
\fi

\paragraph{Procedure.}
The deletion conditions edit the same \paperValue{deletion-start-count/n} original traces with the same editor model as MBT, in the opposite direction.
As in \mbti{}, the editor sees the reasoning section with numbered paragraphs and, in a single call, outputs the sentences to delete with their paragraph numbers and categories. We parse the output, delete the sentences by rule, and clean up the whitespace at the cuts.
The editor is told to keep solution steps, calculations, and the response section.
The eight categories come from research on metacognition (Table~\ref{tab:delete-categories}).
Resuming abandoned paths has no corresponding category, because it adds reasoning that follows from a dropped idea, not a separate metacognitive statement.

\paragraph{Data.}
The deletion run finishes \paperValue{deletion-finished-count/finished} of the \paperValue{deletion-finished-count/total} traces and marks 115,395 deletions, each a sentence or a whole paragraph (\paperValue{deletion-character-percentage/pct}\% of the reasoning characters for Delete-8). Both conditions train on the \paperValue{deletion-train-counts/delete} finished traces among the \paperValue{deletion-train-counts/unedited} of Unedited SFT and keep \paperValue{deletion-token-percentages/d4}\% (Delete-4) and \paperValue{deletion-token-percentages/d8}\% (Delete-8) of their tokens.
The subsampled control draws \paperValue{subsample-count/n} random unedited traces to match the tokens of Delete-8.
It beats Delete-4 by \paperValue{subsample-gain/gain} points with fewer tokens (Table~\ref{tab:deletion}, Figure~\ref{fig:deletion}).

%% file: tables/target_behaviors.tex
\begin{table*}[tp]
\centering
\footnotesize
\setlength{\tabcolsep}{4pt}
\renewcommand{\arraystretch}{1.1}
\begin{tabularx}{\textwidth}{@{}>{\raggedright\arraybackslash}p{2.75cm}>{\raggedright\arraybackslash}X>{\raggedright\arraybackslash}p{4.3cm}r@{}}
\toprule
Target behavior & What it does, and where & Undesirable behaviors it addresses & \makecell[r]{Original\\traces (\%)} \\
\midrule
Subgoal setting & Before solving, states the goal and the subgoals in order, only in traces that start without a plan. & no initial plan & 10.0 \\
Logical check & Right after a deductive step, tests a premise the step rests on, or names a case its split does not cover. & missing logical check, expected-answer guessing, unverified recall & 77.0 \\
Substantive verification & Checks a result by a procedure that would expose an error, such as substitution, small cases, or another derivation. & superficial verification, no substantive verification, unverified conclusion, expected-answer guessing, unverified recall, recall attempt & 96.0 \\
Targeted backtracking & When a path fails, names the earlier point where the problem began and returns there. & shallow backtracking & 2.5 \\
Resuming abandoned paths & Picks up a promising idea that was raised and then dropped without being used. & abandoning promising paths & 76.5 \\
Self-correction & Finds an actual error in an earlier step and appends the correction after it, leaving the step in place. & failure to self-correct, unverified conclusion & 64.0 \\
Sufficiency judgment & Once, where the reasoning ends, consolidates the grounds and settles the answer. & post-answer overthinking, unchecked final answer, expected-answer guessing & 11.0 \\
\bottomrule
\end{tabularx}
\caption{The seven target behaviors. The last column is the share of \paperValue{table-paired/traces} original teacher traces (\paperValue{table-paired/correct} correct, \paperValue{table-paired/wrong} incorrect) that contain the behavior at least once.}
\label{tab:target-behaviors}
\end{table*}

%% file: tables/undesirable_full.tex
\begin{table*}[tp]
\centering
\footnotesize
\setlength{\tabcolsep}{4pt}
\renewcommand{\arraystretch}{1.06}
\begin{tabularx}{\textwidth}{@{}>{\raggedright\arraybackslash}p{4.6cm}>{\raggedright\arraybackslash}Xrr@{}}
\toprule
Behavior & Definition & \makecell[r]{Correct\\traces (\%)} & \makecell[r]{Incorrect\\traces (\%)} \\
\midrule
\tgroup{4}{Undesirable behaviors}
\textbf{Recall attempt} & Tries to recall the source or a known answer of the problem. & 54.3 & 67.4 \\
\textbf{Unverified recall} (sub-behavior of recall attempt) & Asserts a recalled source or answer specifically without checking it. & 18.1 & 34.7 \\
\textbf{Expected-answer guessing} & Chooses an answer by guessing what the problem setter or grader expects instead of using mathematical grounds. & 58.1 & 76.8 \\
Repeating the same interpretation issue & Revisits an interpretation issue, such as the meaning of a symbol, three or more times without new information. & 72.4 & 78.9 \\
Hedged final answer & Gives two or more different answers, or a conditional answer, as the final answer. & 20.0 & 20.0 \\
Repeating the same derivation & Re-derives or rechecks the same result by the same method three or more times without new information. & 84.8 & 87.4 \\
\textbf{Post-answer overthinking} & After reaching an answer, spends at least a quarter of the trace on content not used in the final answer. & 24.8 & 43.2 \\
Rewriting the final answer & Rewrites the final answer three or more times while reasoning. & 75.2 & 83.2 \\
\textbf{Unverified conclusion} & Draws a conclusion from an unchecked calculation or recollection and never checks it. & 22.9 & 32.6 \\
Conclusion contradicting the trace (sub-behavior of unverified conclusion) & The final answer is inconsistent with a result that the trace proved or obtained earlier. & 2.9 & 6.3 \\
Remarks on generation settings & Remarks about generation settings, such as output length, appear in the reasoning. & 75.2 & 81.1 \\
Long manual computation & Performs by hand, over 100 lines or more, a computation or enumeration that a short program would do. & 15.2 & 17.9 \\
Premature stopping & Stops or settles on an answer citing time or effort, without judging that the work is sufficient. & 2.9 & 4.2 \\
Restating the problem & Restates the problem nearly verbatim three or more times. & 50.5 & 60.0 \\
Solving a different problem & Replaces the problem with a different one and answers it. & 4.8 & 1.1 \\
Overgeneralization from small cases & Accepts a general conclusion from a few small cases without proof or a check. & 2.9 & 7.4 \\
\midrule
\tgroup{4}{Desirable behaviors}
\textbf{Substantive verification} & A check that would expose an error if the result were wrong (substitution, small cases, re-derivation by another method). & 95.2 & 94.7 \\
Self-found error correction & Finds an error in an earlier step by itself and corrects it. & 49.5 & 61.1 \\
\textbf{Explicit planning} & Makes an ordered plan before solving. & 17.1 & 9.5 \\
\bottomrule
\end{tabularx}
\caption{All items of the undesirable-behavior annotation scheme on the \paperValue{table-diagnostic/traces} diagnostic teacher traces (\paperValue{table-diagnostic/correct} correct, \paperValue{table-diagnostic/wrong} incorrect). Bold items appear in Figure~\ref{fig:teacher-behaviors}, the last two as 100 minus the rate.}
\label{tab:undesirable-full}
\end{table*}

%% file: figures/teacher_trace.tex
\begin{figure*}[tp]
\centering
\includegraphics[width=\textwidth]{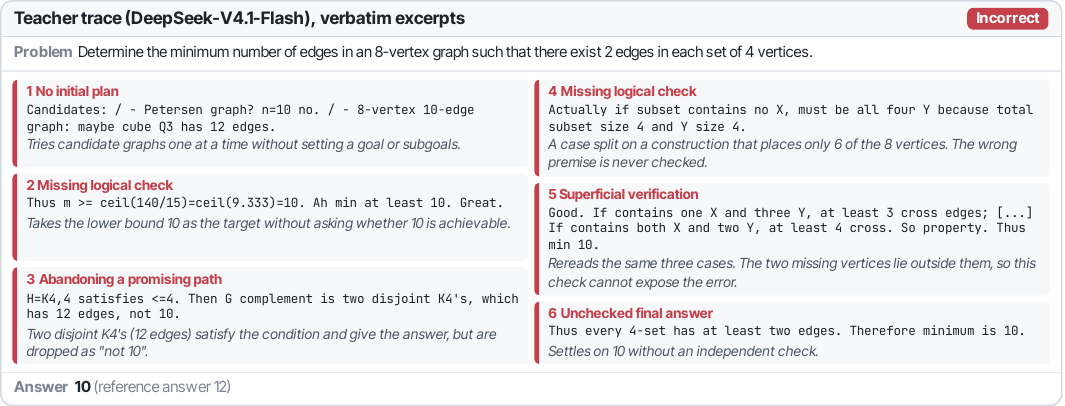}
\caption{A teacher trace in which undesirable behaviors lead to a wrong answer (verbatim excerpts, with labels and notes by the authors). The teacher fixes on the lower bound 10, drops the correct construction, and verifies only by rereading its own case split.}
\label{fig:teacher-trace}
\end{figure*}

%% file: tables/panel_annotator.tex
\centering
\footnotesize
\setlength{\tabcolsep}{4pt}
\begin{tabular}{@{}lc@{}}
\toprule
Comparison & Median kappa \\
\midrule
\tgroup{2}{Agreement with the reference (\paperValue{table-annotator/reference} traces)}
claude-opus-5.5 & \best{0.73} \\
gpt-6.1-sol (used) & \second{0.64} \\
gpt-5.6-luna & 0.55 \\
claude-sonnet-5.5 & 0.54 \\
gpt-6-luna & 0.49 \\
\midrule
\tgroup{2}{Self-consistency of gpt-6.1-sol}
Same \paperValue{table-annotator/repeat} traces annotated twice & 0.89 \\
\bottomrule
\end{tabular}
\caption{Annotator reliability, as the median per-behavior Cohen's kappa under the undesirable-behavior scheme. The agreement of claude-opus-5.5 uses the \paperValue{table-annotator/opus} traces it annotates.}
\label{tab:annotator}

%% file: tables/panel_train-data.tex
\centering
\footnotesize
\setlength{\tabcolsep}{4pt}
\begin{tabular}{@{}lrrr@{}}
\toprule
Condition & Traces & Tokens & vs.\ original \\
\midrule
Unedited SFT & 995 & 22,821,876 & 1.000 \\
\mbta{} & 995 & 24,432,120 & 1.071 \\
\mbti{} & 995 & 23,483,410 & 1.029 \\
\mbtr{} & 773 & 7,557,835 & 0.331 \\
Token-matched control & 1,057 & 24,490,718 & 1.073 \\
ASTRO & 992 & 22,988,634 & 1.007 \\
CoT-Bridge & 995 & 23,055,717 & 1.010 \\
RefAug & 995 & 23,944,165 & 1.049 \\
Gandhi et al. & 987 & 23,931,841 & 1.049 \\
Delete-4 & 993 & 16,814,430 & 0.741 \\
Delete-8 & 993 & 11,337,085 & 0.500 \\
Subsampled control & 510 & 11,337,085 & 0.500 \\
\bottomrule
\end{tabular}
\caption{Training data per condition (tokens under the Qwen3.5-4B chat template). Deletion conditions and the subsampled control are relative to the \paperValue{table-train-data/deletion-traces} traces of the deletion conditions (\paperValue{table-train-data/deletion-tokens} tokens). Counts below \paperValue{table-traces/n} reflect failed edits or generations and traces over \paperValue{table-train-data/max-length} tokens.}
\label{tab:train-data}

%% file: tables/baselines.tex
\begin{table*}[tp]
\centering
\footnotesize
\setlength{\tabcolsep}{4pt}
\renewcommand{\arraystretch}{1.1}
\begin{tabularx}{\textwidth}{@{}>{\raggedright\arraybackslash}p{1.7cm}>{\raggedright\arraybackslash}p{4.6cm}>{\raggedright\arraybackslash}Xc@{}}
\toprule
Method & Original paper & Our implementation & \makecell{Training tokens\\(vs.\ original)} \\
\midrule
ASTRO & Linearizes a search-tree path into one solution, inserting reflection and backtracking phrases where the search returns from an incorrect branch, and a phrase that the solution is on track where it proceeds well. & At the positions where \mbta{} inserts passages, except self-correction positions, inserts the phrases of ASTRO instead. The blank part of a phrase (the point to return to) is filled by DeepSeek-V4.1-Flash. & 1.007 \\
CoT-Bridge & Supplements data with a Qwen2.5-Math-7B model trained to find and fill missing intermediate steps between adjacent steps. & Gives the paper's zero-shot baseline prompt to DeepSeek-V4.1-Flash to find and insert missing steps between paragraphs of the reasoning section. It inserts \paperValue{table-baselines/bridges} steps into \paperValue{table-baselines/bridged-traces} of the \paperValue{table-baselines/bridge-cells} traces (about \paperValue{table-baselines/bridges-per-trace} per edited trace). & 1.010 \\
RefAug & Appends a ``Reflection'' section (an alternative solution and a follow-up problem), written by GPT-4-turbo, which is used only in training. & Gives the paper's prompt to DeepSeek-V4.1-Flash to append a reflection section after the response section. The reasoning section is unchanged, and at evaluation the output is cut before ``Reflection:''. & 1.049 \\
Gandhi et al. & Has Claude 3.5 Sonnet solve problems anew with a system prompt asking for the four cognitive behaviors, without filtering by correctness. & Has the teacher solve each problem again with the system prompt from the paper's repository adapted to math. The line describing the Countdown task is replaced by the math problem, and the answer-format instruction and the HINT line are removed, while the behavior instructions and five examples are kept. New answers are used without filtering, and correctness under math\_verify changes for \paperValue{table-baselines/gandhi-changed} of \paperValue{table-baselines/gandhi-traces} problems (\paperValue{table-baselines/gandhi-changed-pct}\%). & 1.049 \\
TI (Thinking Intervention) & Without training, inserts a first-person instruction where reasoning begins and lets the model continue. & At evaluation, inserts a first-person sentence asking for the behaviors at the start of the Base model's thinking, e.g., ``Okay, the user asked for this. I should ... plan the steps, ... catch and correct any mistake, ...'' & none \\
\bottomrule
\end{tabularx}
\caption{Reimplementation of prior methods. The four training-based methods use the same problems, students, and training settings as MBT, and all except Gandhi et al.\ start from the same original traces.}
\label{tab:baselines}
\end{table*}

%% file: tables/panel_editing-full.tex
\centering
\footnotesize
\setlength{\tabcolsep}{1.8pt}
\begin{tabular}{@{}lccc@{}}
\toprule
 & \mbtr{} & \mbti{} & \mbta{} \\
\midrule
\tgroup{4}{Completion}
Edited successfully (of \paperValue{table-traces/n}) & 776 & 1,000 & 1,000 \\
Attempts per trace (mean) & 4.24 & 1.01 & 1.00 \\
\midrule
\tgroup{4}{Length ratio $\rho$}
Median & 1.000 & 1.048 & 1.112 \\
Mean & 0.864 & 1.138 & 1.271 \\
Max & 3.987 & 3.638 & 5.707 \\
Below 0.5 (\%) & 28.1 & 0.0 & 0.0 \\
Median, originals $\geq$32k tokens & 0.078 & 1.012 & 1.040 \\
Training tokens (vs.\ original) & 0.331 & 1.029 & 1.071 \\
\midrule
\tgroup{4}{Insertion position}
Insertions (total) & 4,467 & 6,472 & 10,759 \\
Insertions per trace (mean) & 5.8 & 6.5 & 10.8 \\
Median position & 0.150 & 0.246 & 0.356 \\
In the first 10\% (\%) & 41.5 & 31.9 & 23.4 \\
KS statistic $D$ (lower is more even) & 0.363 & \second{0.257} & \best{0.161} \\
\bottomrule
\end{tabular}
\caption{Editing statistics on the same \paperValue{table-traces/n} teacher traces. Length statistics of \mbtr{} use its \paperValue{table-editing/rewrite-edited} successful edits, and positions are relative to the original reasoning section, from 0 (start) to 1 (end).}
\label{tab:editing-full}

%% file: tables/panel_mbtr-failures.tex
\centering
\footnotesize
\setlength{\tabcolsep}{4pt}
\begin{tabularx}{\linewidth}{@{}>{\raggedright\arraybackslash}Xrr@{}}
\toprule
Failure reason (last call) & Traces & Share \\
\midrule
No section headers (solves the problem instead of returning the file) & 156 & 69.6\% \\
Output limit hit five times in a row & 44 & 19.6\% \\
Copies the instructions after the last line of the file & 13 & 5.8\% \\
Wrong section order or a duplicated section & 5 & 2.2\% \\
Replaces original text with an omission phrase & 5 & 2.2\% \\
Request timeout & 1 & 0.4\% \\
\bottomrule
\end{tabularx}
\caption{Why \mbtr{} fails on 224 traces.}
\label{tab:mbtr-failures}

%% file: tables/panel_mbta-cost.tex
\centering
\footnotesize
\setlength{\tabcolsep}{1.9pt}
\begin{tabular}{@{}lrrrr@{}}
\toprule
Quantity & Median & Mean & Max & Total \\
\midrule
Commits & 15 & 17.0 & 54 & 16,993 \\
Agent turns & 52 & 56.4 & 169 & 56,401 \\
Time (s) & 872 & 888 & 2,380 & 887,964 \\
Cost (USD) & 0.147 & 0.149 & 0.392 & 149.26 \\
Input tokens, cache miss & 87.3k & 97.7k & 330.6k & 97.7M \\
Input tokens, cache hit & 8.89M & 10.86M & 52.17M & 10.86B \\
Output tokens & 167.5k & 170.1k & 476.9k & 170.1M \\
\bottomrule
\end{tabular}
\caption{Work of \mbta{} per trace over \paperValue{table-traces/n} traces. An agent turn is one call of the agent to the editor model, and cost is at the official DeepSeek API prices at the time of the runs.}
\label{tab:mbta-cost}

%% file: tables/panel_fix-commits.tex
\centering
\footnotesize
\setlength{\tabcolsep}{4pt}
\begin{tabular}{@{}lrr@{}}
\toprule
Type of repair & Commits & Share \\
\midrule
Sentence polishing & 3,311 & 82.4\% \\
Content correction & 397 & 9.9\% \\
Notation & 203 & 5.1\% \\
Relocation & 107 & 2.7\% \\
\bottomrule
\end{tabular}
\caption{Commits in which the \mbta{} agent repairs its own earlier insertions (4,018 commits in \paperValue{table-repairs/traces} traces), classified by keywords in the commit subjects.}
\label{tab:fix-commits}

%% file: tables/panel_epochs.tex
\centering
\footnotesize
\setlength{\tabcolsep}{4pt}
\begin{tabular}{@{}lccc@{}}
\toprule
Condition & ep5 & ep10 & ep15 \\
\midrule
Unedited SFT & \second{75.7} & 82.8 & 83.1 \\
\mbta{} & 72.9 & \best{84.8} & \best{85.4} \\
Token-matched control & 74.8 & 83.0 & 83.2 \\
\mbti{} & 74.3 & \second{83.7} & 83.8 \\
\mbtr{} & 64.1 & 76.0 & 73.9 \\
ASTRO & 72.8 & \second{83.7} & 81.8 \\
CoT-Bridge & 73.4 & 83.5 & 83.2 \\
RefAug & 74.0 & 82.7 & 82.9 \\
Gandhi et al. & \best{76.1} & 82.5 & \second{84.2} \\
Delete-4 & 61.5 & 68.6 & 66.7 \\
Delete-8 & 50.7 & 58.2 & 54.7 \\
Subsampled control & 38.6 & 77.4 & 83.6 \\
\bottomrule
\end{tabular}
\caption{Math avg@8 (\%) on the \paperValue{table-math/problems} problems per epoch. Unedited SFT and \mbta{} are the runs of Table~\ref{tab:main}, and the others are single runs. Bold and underline mark the best and second best per column.}
\label{tab:epochs}

%% file: tables/panel_other-students.tex
\centering
\footnotesize
\setlength{\tabcolsep}{2.3pt}
\begin{tabular}{@{}lcccccc@{}}
\toprule
 & \multicolumn{2}{c}{AIME} & \multicolumn{3}{c}{HMMT} & \\
\cmidrule(lr){2-3}\cmidrule(lr){4-6}
Model / condition & 25 & 26 & Feb25 & Feb26 & Nov25 & Avg. \\
\midrule
\tgroup{7}{MiniCPM5-2B-Base}
\quad Base model & 27.5 & 31.2 & 17.5 & 21.2 & 23.3 & 24.1 \\
\quad Unedited SFT & \best{41.2} & 34.6 & 20.8 & 23.1 & 23.3 & 28.5 \\
\quad \mbta{} & 37.5 & \best{37.9} & \best{27.9} & \best{25.4} & \best{28.8} & \best{31.4} \\
\midrule
\tgroup{7}{Granite 4.1 3B Base}
\quad Base model$^{a}$ & 8.8 & 8.8 & 4.6 & 5.7 & 2.9 & 6.1 \\
\quad Unedited SFT & 32.9 & 32.1 & 23.3 & \best{27.7} & \best{28.3} & 28.8 \\
\quad \mbta{} & \best{34.6} & \best{36.7} & \best{25.8} & 27.3 & 27.9 & \best{30.4} \\
\bottomrule
\end{tabular}
\caption{Competition math (avg@8, \%) for the other students, trained with the same data and settings. Bold marks the better of Unedited SFT and \mbta{}. $^{a}$Queried in plain text without a chat template.}
\label{tab:other-students}

%% file: tables/panel_seeds.tex
\centering
\footnotesize
\setlength{\tabcolsep}{4pt}
\begin{tabular}{@{}llccc@{}}
\toprule
Condition & Run & ep5 & ep10 & ep15 \\
\midrule
Unedited SFT & run 1$^{*}$ & \best{75.7} & 82.8 & 83.1 \\
 & run 2 & 74.8 & 82.8 & 83.7 \\
 & run 3 & \second{75.2} & 82.8 & 82.8 \\
\midrule
\mbta{} & run 1$^{*}$ & 72.9 & \best{84.8} & \best{85.4} \\
 & run 2 & 71.9 & \best{84.8} & 85.0 \\
 & run 3 & 71.6 & 83.4 & \best{85.4} \\
\bottomrule
\end{tabular}
\caption{Three runs with different seeds (math avg@8, \%). $^{*}$Runs reported in the main text. Bold and underline mark the best and second best per column.}
\label{tab:seeds}

%% file: tables/clogit.tex
\begin{table}[tp]
\centering
\footnotesize
\setlength{\tabcolsep}{1.8pt}
\begin{tabular}{@{}>{\raggedright\arraybackslash}p{1.95cm}cccc@{}}
\toprule
Behavior & \makecell{Odds ratio\\{[95\% CI]}} & \makecell{Raw\\$p$} & \makecell{Holm\\$p$} & \makecell{Discordant\\pairs} \\
\midrule
Subgoal setting & 0.89 [0.62, 1.29] & 0.53 & 1 & 116 \\
\textbf{Logical check} & \textbf{2.60 [1.38, 4.91]} & \textbf{0.0031} & \textbf{0.022} & \textbf{50} \\
Substantive verification & 7.62 [1.30, 44.72] & 0.024 & 0.15 & 14 \\
Targeted backtracking & 1.07 [0.74, 1.54] & 0.74 & 1 & 116 \\
Resuming abandoned paths & 0.93 [0.65, 1.32] & 0.67 & 1 & 133 \\
Self-correction & 1.31 [0.65, 2.63] & 0.45 & 1 & 39 \\
Sufficiency judgment & 1.24 [0.81, 1.89] & 0.32 & 1 & 88 \\
\bottomrule
\end{tabular}
\caption{Association of each behavior with correctness within the same problem and condition (437 pairs). Bold marks Holm-adjusted $p < 0.05$.}
\label{tab:clogit}
\end{table}

%% file: tables/panel_deletion.tex
\centering
\footnotesize
\setlength{\tabcolsep}{3pt}
\begin{tabular}{@{}lcc@{}}
\toprule
Condition & Tokens & Score \\
\midrule
Base model (no training) &  & 77.8 \\
Unedited SFT & 1.000 & \second{83.1} \\
Delete-4 & 0.741 & 66.7 \\
Delete-8 & 0.500 & 54.7 \\
Subsampled control & 0.500 & \best{83.6} \\
\bottomrule
\end{tabular}
\caption{Deletion results (math avg@8, \%). Tokens are relative to the original tokens of the same traces. Bold and underline mark the two best trained conditions.}
\label{tab:deletion}